\documentclass[10pt,twocolumn,letterpaper]{article}

\usepackage{cvpr}              

\usepackage{graphicx}
\usepackage{amsmath}
\usepackage{amssymb}
\usepackage{booktabs}
\usepackage{graphbox,xcolor}
\definecolor{darkgoldenrod}{rgb}{0.72, 0.53, 0.04}
\definecolor{darkgreen}{rgb}{0.0, 0.2, 0.13}

\usepackage[pagebackref,breaklinks,colorlinks]{hyperref}

\usepackage[capitalize]{cleveref}
\crefname{section}{Sec.}{Secs.}
\Crefname{section}{Section}{Sections}
\Crefname{table}{Table}{Tables}
\crefname{table}{Tab.}{Tabs.}
\newcommand{\norm}[1]{\left\lVert#1\right\rVert}
\usepackage{color,soul}
\usepackage{tabstackengine}
\usepackage{booktabs}

\def\cvprPaperID{7612}
\def\confName{CVPR}
\def\confYear{2022}

\begin{document}

\title{ZeroCap: Zero-Shot Image-to-Text Generation for Visual-Semantic Arithmetic}

\author{Yoad Tewel \quad\quad\quad Yoav Shalev\quad\quad\quad Idan Schwartz \quad\quad\quad Lior Wolf\\
School of Computer Science, Tel Aviv University
}
\maketitle

\begin{abstract}

Recent text-to-image matching models apply contrastive learning to large corpora of uncurated pairs of images and sentences. While such models can provide a powerful score for matching and subsequent zero-shot tasks, they are not capable of generating caption given an image. In this work, we repurpose such models to generate a descriptive text given an image at inference time, without any further training or tuning step. This is done by combining the visual-semantic model with a large language model, benefiting from the knowledge in both web-scale models. The resulting captions are much less restrictive than those obtained by supervised captioning methods. Moreover, as a zero-shot learning method, it is extremely flexible and we demonstrate its ability to perform image arithmetic in which the inputs can be either images or text and the output is a sentence. This enables novel high-level vision capabilities such as comparing two images or solving visual analogy tests.
Our code is available at: \url{https://github.com/YoadTew/zero-shot-image-to-text}.



\end{abstract}

\section{Introduction}
\label{sec:intro}

Deep learning has led to at least three major revolutions in computer vision: (i) machines that achieve, in multiple domains, what is considered a human level of performance earlier than anticipated~\cite{krizhevsky2012imagenet,deepface}, (ii) effective transfer learning, which supports rapid modeling of new domains~\cite{yosinski2014transferable}, and (iii) a leap in unsupervised learning through the use of adversarial and self-supervised learning~\cite{GAN,chen2020simple}.

A fourth revolution that is currently taking place is that of zero-shot learning. A seminal work by OpenAI presented the transformer-based~\cite{vaswani2017attention} GPT-3 model~\cite{brown2020language}. This model is trained on extremely large text corpora and can then generate text given a prompt. If the prompt contains an instruction, GTP-3 can often carry it out. For example, given the prompt ``Translate English to French: typical $\rightarrow$ typique, house $\rightarrow \dots$ '' would generate the word ``maison.'' 

Impressive zero-shot capability was later on demonstrated, also by OpenAI, in computer vision. While state-of-the-art computer vision models are often trained as task-specific models that infer a fixed number of labels, Radford~\etal\cite{radford2021learning} have presented the CLIP image-text transformer model, which can perform tens of downstream tasks, without further training, with an accuracy comparable to the state of the art. This is done by selecting, given an image, the best match out of sentences of the form ``This is an image of X.'' Subsequently, Ramesh~\etal\cite{ramesh2021zero} presented a bi-modal Transformer termed DALL-E, which generates images that match a given description in unseen domains with unprecedented performance.

In this work, we employ CLIP to perform the inverse task of DALL-E, namely zero-shot image captioning. Given an image, we employ CLIP together with the GPT-2 language model~\cite{radford2019language} (we do not have access to GPT-3) to generate a textual description of the input image. This adds a new image-analysis capability to CLIP, beyond the fixed-prompt zero-shot learning demonstrated by Radford~\etal

As a zero-shot method, our approach does not involve any training. One can argue that the underlying CLIP model is trained with exactly the same type of supervision that image captioning methods~\cite{zhang2021vinvl,shen2021much} are trained on, i.e., pairs of matching images and captions. However, image captioning methods are trained from curated sources, such as MS-COCO~\cite{ty2014coco} or Visual Genome~\cite{krishnavisualgenome}, while CLIP is trained on WebImageText (WIT), which is an automatically collected web-scale dataset. Previous attempts to train a captioning model on WIT have led to poor performance in recognizing the objects in the image, see Sec.~\ref{sec:related}.

 \begin{figure*}[t]
	\centering
    \includegraphics[width=0.95\linewidth]{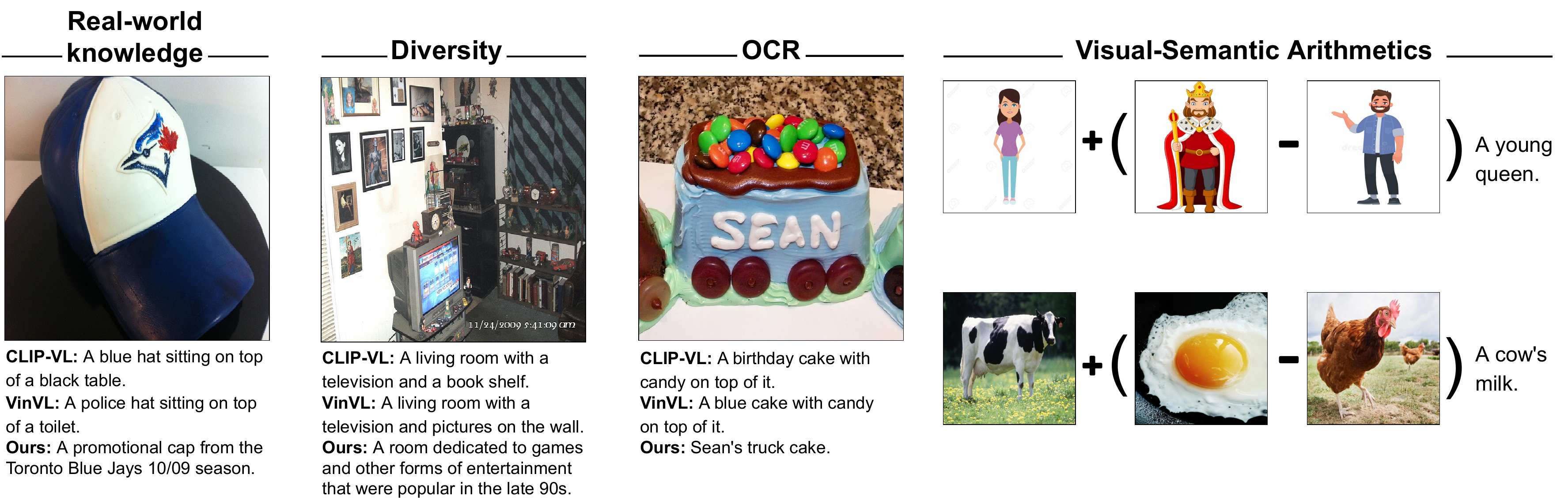}
    \caption{
    Our novel captioning method ZeroCap exhibits real-world knowledge, generates text that is more diverse and less scripted than existing methods, can address the written content of an image, and can perform visual-semantic arithmetic. 
    }\vspace{-15pt}
    \label{fig:teaser}
\end{figure*}

As a result of the difference in both methodology and underlying data, the captions produced by our method are very different from those obtained by the supervised captioning methods. While supervised methods can mimic human annotators and provide similar sentences, in terms of conventional NLP metrics (such as BLEU~\cite{papineni2002bleu}) to the ground truth sentences, our results exhibit much more freedom and match the image better in the visual-semantic CLIP embedding space (ours is optimized for this). Moreover, the semantic knowledge incorporated into CLIP and GPT-2 is manifested in the resulting caption, see \cref{fig:teaser}.

In addition to the different nature of the obtained captions, our method is also more flexible, since all the computing occurs at inference time. Specifically, we show the ability to perform semantic analysis in image space by using a new form of arithmetic. A well-known example for concept arithmetic in NLP is that of retrieving the word `queen' as the closest word, in the embedding space, to the equation involving the embedding vectors associated with `king,' `man,' and `woman,' after subtracting the 2nd from the 1st and adding the 3rd. We present the novel ability to do the same, only with images instead of words, such that the result is generated as a short sentences, and not just a word, see \cref{fig:teaser}.

As a corollary, we can, for example, ask what the difference is between two scenes. This ability to compare two images semantically is a novel computer vision capability, which further demonstrates the power of zero-shot learning.

\section{Related work}
\label{sec:related}

 The first deep captioning methods applied RNNs to generate sequences of words~\cite{Mao2014DeepCW,klein2014fisher}. Attention was added to identify relevant salient objects~\cite{xu2015show, schwartz2017high, schwartz2019factor}. 
 Graph neural networks and scene graphs incorporated spatial as well as semantic relationships between objects~\cite{yao2018exploring, kipf2016semi, yang2019auto}. Subsequently, Transformers modeled interactions among all image elements with self-attention~\cite{vaswani2017attention, pan2020x, dosovitskiy2021image, Schwartz_2019_CVPR}. On the text modeling side of the problem, language models (LMs) have also advanced with the development of LSTMs~\cite{vinyals2015show,donahue2015long}, CNNs~\cite{aneja2018convolutional} and Transformers~\cite{luo2021dual,guo2020normalized, herdade2019image}. Language improvements include devising better image grounding~\cite{lu2018neural}, decoding non-visual words (e.g., `the,' `and')~\cite{lu2017knowing}, generating fine, novel and diverse sentences~\cite{gu2018stack, chatterjee2018diverse, wang2017diverse}, and incorporating information from different semantic taggers~\cite{cornia2020meshed, ji2021improving}.

In recent years, significant improvements have been achieved by utilizing large-scale vision-language data sets. The unsupervised data is used as a pre-training phase, to initialize models with image-text correspondence~\cite{zhang2021vinvl, devlin2018bert, li2020oscar}. With this technique, millions of image and text pairs from the web can be adopted. Nevertheless, in previous work we are aware of, all captioning models employ human-annotated datasets, such as MS-COCO or the Visual Genome, in the last stage of training. 





 It is likely impossible to construct a database of curated captions that is large enough, to describe even a modestly large fraction of plausible images and objects. This results in biases~\cite{verma2021removing, gat2020removing, gat2021perceptual}. Several approaches focused on describing novel objects by conditioning the model on external unsupervised data during training~\cite{hendricks2016deep,venugopalan2017captioning, agrawal2019nocaps}. Alternatively, external object taggers can be used during different phases (e.g., pre-training, training, or inference)~\cite{li2019pointing, lu2018neural, feng2020cascaded, anderson2016guided, hu2021vivo}. Semi-supervised methods are also available~\cite{kim2019image}.  Unsupervised approaches can be achieved by training with a visual concept detector or by learning a joint image-language embedding space~\cite{feng2019unsupervised, laina2019towards}. 
 In contrast, our method makes use of an existing image-text alignment score to direct an existing large-scale LM toward a given image without training.

CLIP is trained on 400M images/sentence pairs from the web~\cite{radford2021learning}, resulting with a powerful text-image matching score. Originally CLIP's authors explored training an image-to-caption language model with this training set, but found that it struggled with zero-shot transfer. In a 16 GPU-day experiment, a language model only achieved 16\% accuracy on ImageNet~\cite{imagenet_cvpr09}. CLIP achieves the same level of accuracy roughly 10x faster.

Using prompts, it is possible to imitate some capabilities of text generation. For example, CLIP-based applications exhibit zero-shot solving capabilities in various scenarios never seen before. With careful engineering of the prompt, one can, for example, improve detection of unseen objects~\cite{gu2021zero}.  Zero-shot prompt engineering has also been used for higher-level tasks (e.g., VQA), but it is nowhere near the level of supervised methods~\cite{shen2021much}. 

CLIP also provides powerful means for supporting text-driven image manipulation with Generative Adversarial Networks (GANs) or other generative models~\cite{patashnik2021styleclip, sanghi2021clip, chefer2021image}. Our work explores the other direction: generating text using an image, by guiding a large-scale LM with CLIP.

Guided language modeling has become a primary challenge, as researchers strive to tune prior knowledge within large-scale LMs, such as GPT-2~\cite{radford2019language}.  Fine-tuning is often accomplished by employing Reinforcement Learning~\cite{ziegler2019finetuning} or GANs~\cite{yu2016seqgan} for each attribute separately. Disentangling the latent representations into style and content is also relevant in terms of text style transfer~\cite{shen2017style, hu2017toward}. A controllable LM can also be formed using fixed control codes~\cite{keskar2019ctrl}. Ideally, conditioning should be applied directly to the existing large-scale LM, without the need for fine-tuning. Several studies have explored the idea of steering an LM using small neural networks~\cite{gu2016learning, chen2018stable}. Following that, PPLM~\cite{Dathathri2020Plug} demonstrated that a simple attribute classifier could steer a model without any further training. With our work, we present a novel visual LM guidance from visual cues.

\begin{figure*}[t]
	\centering
    \includegraphics[width=.80\linewidth]{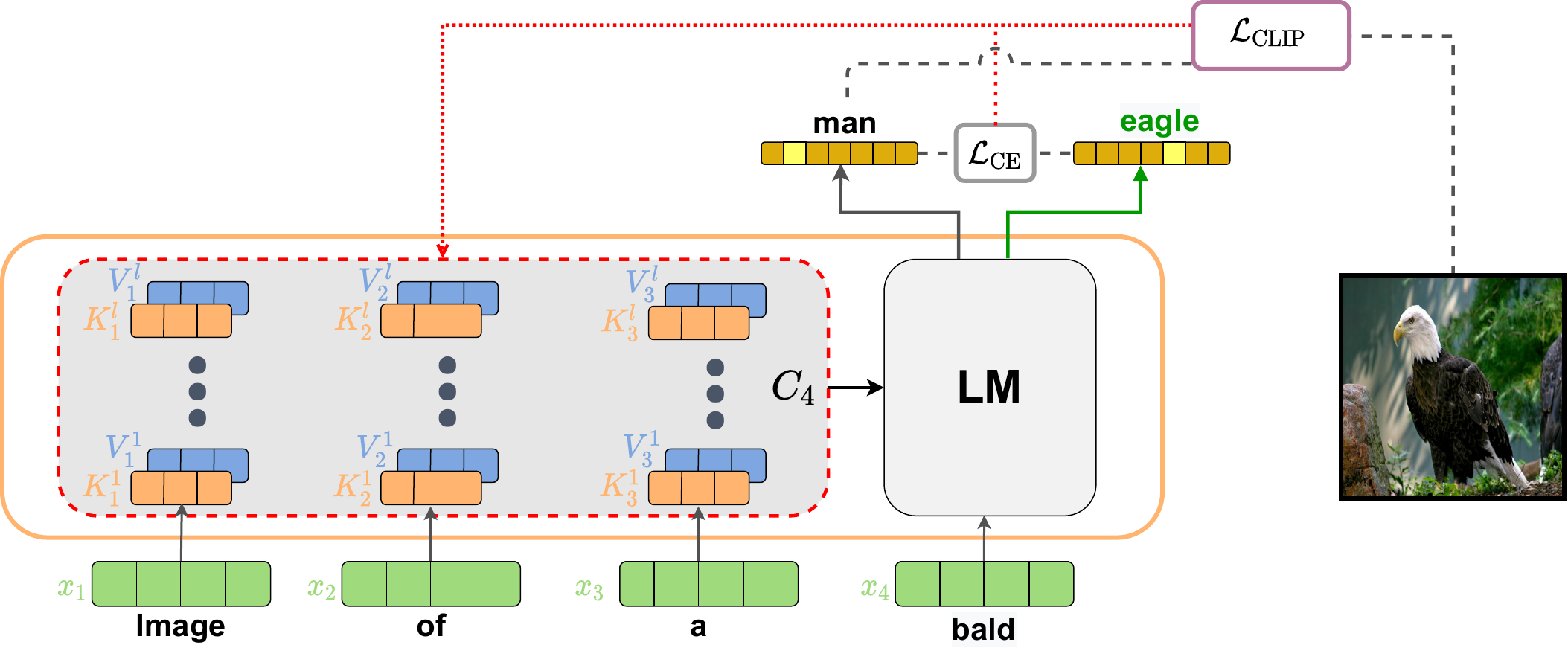}
    \caption{An overview of our approach. We guide the model towards the phrase `eagle' instead of `man'. We do this by adjusting the context ($C_4$), using the gradients of CLIP loss ($\mathcal{L}_\text{CLIP}$) illustrated with a red arrow. To maintain language attributes, we optimize the minimum distance to the original distribution of the LM  ($\mathcal{L}_\text{CE}$).} \vspace{-15pt}
    \label{fig:approach}
\end{figure*}

\section{Method}
\label{sec:method}
Visual captioning is the process of generating a descriptive sentence for an image. It can be formalized as a sequence generation problem given an input image $I$, i.e., as a conditional probability inference for the $i$-th word $x_i$ of the sentence, i.e.,  $p(x_i|[x_{t}]_{t<i}, I)$. 

This is typically accomplished in a supervised manner, by optimizing weights to reproduce ground truth sentences. However, since carefully curated datasets are small, and cannot adequately describe all images, the sentences generated often describe the content at the basic level of the objects present in the scene and sound artificial. 
Such problems can be mitigated with the use of web-scale datasets. 
We present a zero-shot method for guiding large-scale language models with a large-scale text-image alignment model.


\noindent{\bf Overview\quad} Our approach uses a transformer-based LM (e.g., GPT-2) to infer the next word from an initial prompt, such as ``Image of a,'' as illustrated in \cref{fig:approach}. To incorporate image-related knowledge to the auto-regression process, a calibrated CLIP loss $\mathcal{L}_{\text{CLIP}}$ stimulates the model to generate sentences that describe a given image. An additional loss term $\mathcal{L}_{\text{CE}}$  is used to maintain the next token distribution similar to the original language model. Optimization occurs during auto-regression, and repeated for each token. 

Furthermore, the flexibility of our method enables the capturing of semantic relations through simple arithmetic of visual cues in CLIP's embedding space. Finally, combining multi-modal encoders with our method allows knowledge to be extracted in a new way that mixes between text and images.

.


\noindent{\bf Language models\quad} In recent years, LMs have improved significantly and are getting closer to AI-complete capabilities, including broad external knowledge and solving a wide variety of tasks with limited supervision. A Transformer-based LM typically models interactions between the generated token and past tokens at each time-step. 

Recall that the transformer block has three embedding functions $K, Q, V$~\cite{vaswani2017attention}. The first two, $K, Q$, learn the token interactions that determine the distribution over $V$. The attention mechanism pools values based on the similarity between queries and keys. Specifically, the pooled value for each token $i$ depends on the query associated with this token $Q_i$, which is computed using the function $Q$ over the current embedding of this token. The result is obtained as the weighted average of the value vectors, based on the cosine similarity between $Q_i$ and the keys associated with all tokens $K_j$. 

While $K$ and $V$ are functions, the obtained key and values $K_j$ and $V_j$ are used repeatedly when generating text, one word at a time. $K_j$ and $V_j$ can therefore be stored in what is called a context cache, in order to keep track of past embedding outputs of $K$ and $V$. The sequence generation process can then be written as
\begin{equation}
\label{eq:lm}
x_{i+1} = \operatorname{LM}\left( x_i, [(K_j^l, V_j^l)]_{j<i,1\leq l\leq L} \right),
\end{equation}
where $x_i$ is the $i$-th word of the generated sentence, $K_j^l, V_j^l$ are the context transformer's key and value of the $j$-th token, and $l$ indicates the index of the transformer layers, out of a total of $L$ layers. Our method employs GPT-2, which has $L=24$ layers.  

We next describe how we align our LM with the input image. We do so by modifying, during inference, the values of the context cache $C_i = [(K_j^l, V_j^l)]_{j<i,1\leq l\leq L}$ leaving the LM unchanged.



\smallskip
\noindent{\bf CLIP-Guided language modelling\quad}
Our goal is to guide the LM towards a desired visual direction with each generation step. The guidance we propose has two primary goals: (i) alignment with the given image; and (ii) maintaining language attributes. The first goal is obtained through CLIP, which is used to assess the relatedness of a token to an image and adjust the model (or, rather, the cache) accordingly. For the second goal, we regularize the objective to be similar to the original target output, i.e., before it was modified. 

The solved optimization problem adjusts the context cache $C_i$ at each time point and is formally defined as %
$    \operatorname*{arg\,min}_{C_i} \mathcal{L}_{\text{CLIP}}( \operatorname{LM}\left( x_i, C_i \right), I)$
\begin{equation}
\label{eq:loss}
 + \lambda \mathcal{L}_{\operatorname{CE}}\left(\operatorname{LM}\left( x_i, C_i \right), \hat x_{i+1}\right),
\end{equation}
where $\hat x_{i+1}$  is the token distribution obtained using the original, unmodified, context cache. The second term employs CE loss to ensure that the probability distribution across words with the modified context is close to the one of the original LM.  The hyperparameter $\lambda$  balances  the two loss terms. It was set to 0.2 early on in the development process and was unmodified since. Next, we explain how the CLIP loss term is calculated.  

\noindent{\bf CLIP loss\quad} We calculate image relevance for the possible tokens at time $i$. It is sufficient to compute potentials for the top 512 token candidates and set the rest to zero potential for efficiency.  To this end, the corresponding candidate sentence $s_i^k = (x_1, ..., x_{i-1},x_i^k)$ for the $k$-th candidate token is matched against the image $I$.  

The clip potential of the $k$-th token is computed as
\begin{equation}\label{eq:clip-poten}
    p_i^k \propto \exp(\operatorname{D}_{\text{CLIP}}(E_{\text{Text}}(s_i^k), E_{\text{Image}}(I)) / \tau_c)),
\end{equation}
where $D_{\text{CLIP}}$ is the cosine distance between CLIP's embeddings of the text (i.e., $E_{\text{Text}}$) and the image (i.e., $E_{\text{Image}}$), and $\tau_c > 0$ is a temperature hyperparameter that controls the sharpness of the target distribution. In all our experiments, we set $\tau_c$ to 0.01.

The CLIP loss is defined as the cross-entropy loss between the clip potential distribution and the target distribution of the next token $x_{i+1}$ obtained by the language model:
\begin{equation}
    \mathcal{L}_{\text{CLIP}} = \operatorname{CE}(p_i, x_{i+1}).
\end{equation}
This loss encourages words that lead to higher CLIP matching scores between the image and the generated sentence.

\noindent{\bf Inference\quad}
As a zero-shot method, no training takes place. At inference time one optimizes the problem in \cref{eq:loss}, which we denote as $p(x_{i+1}|C_i)$, by conducting five steps of gradient descent, i.e.,
\begin{equation}\label{eq:gd}
   C_i \longleftarrow C_i + \alpha \frac{\nabla_{C_i} p(x_{i+1}|C_i)}{\norm{\nabla_{C_i} p(x_{i+1}|C_i)}^2}.
\end{equation}
This update rule is simplified for brevity. With each newly-generated token, the optimization is re-done. In our implementation, the gradients are normalized with Euclidean normalization before each step, separately for each transformer layer. We set the learning rate $\alpha$ to 0.3. 

\noindent{\bf Beam search\quad} The byte-level tokenizer used employs 256 bytes of base tokens to represent every word in existence~\cite{radford2019language}. Any word can also be split into more than one subwords, e.g., the word `zebra' is tokenized as `zeb' and `ra'. As a result, we found that images of zebras are described as striped animals, since the token `zeb' is not picked. Beam search inference helps solve this problem by enabling the search to be conducted in a less myopic way.

\begin{table*}[t]
    \centering

           \begin{tabular}{lcccccccc} 
                \toprule
                &  \multicolumn{5}{c}{\textbf{Supervised Metrics}} & \multicolumn{2}{c}{\textbf{Diversity Metrics}} &  \multicolumn{1}{c}{\textbf{Unsupervised Metric}}  \\
                \cmidrule(lr){2-6} 	\cmidrule(lr){7-8} 	\cmidrule(lr){9-9} 	
                Method                    & B@4  & M  & C & S & $\text{CLIP-S}^{\text{Ref}}$ & Vocab & \%Novel & CLIP-S \\  
                \midrule
                ClipCap \cite{clip-prefix} & 32.15 & 27.1 & 108.35 & 20.12 & 0.81& 1650& 66.4\% & 0.77 \\
                CLIP-VL~\cite{shen2021much} & 40.2 &29.7  & 134.2 & 23.8 & 0.82& 2464& 85.1\% & 0.77   \\
                VinVL~\cite{zhang2021vinvl} & {\bf 41.0} & {\bf 31.1}& {\bf 140.9} & {\bf 25.2} & {\bf 0.83}& 1125 & 77.9\% & 0.78 \\
                \midrule
                Ours & 2.6 &11.5 & 14.6 & 5.5 & 0.79& {\bf8681} & \textbf{100\%} & \textbf{0.87} \\
               \bottomrule
            \end{tabular}\vspace{-5pt}
        \caption{For each method, we report supervised metrics (i.e., ones requiring human references): B@1 = BLEU-1, M = METEOR, C = CIDEr, S = SPICE. We also report diversity metrics, which measures the vocabulary size (Vocab), and the number of novel sentences w.r.t the training set (\%Novel). Finally, we report semantic relatedness to the image (CLIP-S), and to the human references ($\text{CLIP-S}^{\text{Ref}}$) based on CLIP's embeddings.}\vspace{-10pt}
     \label{tab:results}
    \end{table*}

 \section{Visual-Semantic Arithmetic}

 Recent studies suggested that CLIP multi-modal representation holds an elaborate concept taxonomy~\cite{goh2021multimodal}. In accordance with this intuition, we find that our method can express CLIP's embedding in a textual way. For instance, subtracting between CLIP-encoded images and applying our method transcribes a relationship between the two images. 
 Furthermore, 
 by summing vectors we can steer the generated caption towards a conceptual direction. 
 
 To perform arithmetic in CLIP's embedding space, we first encode the image/text using CLIP's image/text encoder. For instance, let $I_1, I_2$ be two images. We encode the images with CLIP's encoder, i.e., $E_{\text{image}}(I_1), E_{\text{image}}(I_2)$. Next, we carry out the desired arithmetic, e.g., addition with $E_{\text{image}}(I_1)+E_{\text{image}}(I_2)$. Finally, we use the obtained result instead of the image encoding $E_\text{image}(I)$ within \cref{eq:clip-poten} to steer the generated sentence.
 
 Consequently, we can generate detailed knowledge of the external world by moving in conceptual directions. This way, our method can answer questions expressed visually, for example, ``who is the president of Germany?'' To achieve this, we subtract ``America's flag'' from an image of ``Obama'' and obtain a presidential-direction, to which we can then add the image of a second country's flag. 
 
Our approach extends beyond visual interactions alone. Using CLIP's textual encoder, interaction with a natural language is possible. In this case, 
 one performs arithmetic operations in the embedding space such that the expression contains both image- and text-embeddings.



\section{Experiments}


For all the results reported in this section, we used a strategy for reducing repetitions, in which the probability for generating tokens that were generated at the last four time-steps was decreased by a factor of two. We also incorporated a mechanism that directly controls the length of the generated text by multiplying the probability of the end token by a factor of $f_e$, starting from time-step $t_e$. We use $f_e=1.04$ and $t_e=3$ for image captioning, and $f_e=1.06$ and $t_e=1$ for image arithmetic. On a single Titan X GPU, five beams and 512 candidate tokens can be generated in three seconds. Inference time is proportional to the number of candidates and beams.

\subsection{Image Captioning}

We begin by studying our zero-shot method for caption generation. Notably, we find our captions to exhibit human-like characteristics, such as generating diverse captions, reading, exploiting a wide range of external knowledge, and coping with abstract concepts. 
In \cref{tab:results}, we present our results for COCO's test set~\cite{ty2014coco}. Two recent baselines that use CLIP's embedding are compared to: ClipCap~\cite{clip-prefix} and  CLIP-VL~\cite{shen2021much}. In ClipCap,  the image is encoded using CLIP and the representation is transferred and plugged as a token into a fine-tuned GPT-2. 
CLIP-VL incorporates spatial grid features from CLIP into a transformer network. Another method, VinVL~\cite{zhang2021vinvl} is a state-of-the-art technique. 

We first consider supervised metrics, i.e., metrics requiring human references. These metrics include the BLEU~\cite{Papineni2001BleuAM}, METEOR~\cite{Banerjee2005METEORAA}, CIDEr~\cite{Vedantam2014CIDErCI}, SPICE~\cite{anderson2016spice}, and CLIPScoreRef that we discuss below. As can be seen, our method lags in these metrics in comparison to the supervised captioning methods. Since the ground truth human annotation is obtained similarly to the training set, with the same group of annotators using similar terms, there is a clear advantage for methods trained on COCO annotations. 

We next consider diversity metrics. Our vocabulary over COCO's test set is significantly larger than previous approaches (8681 vs. 2464). In addition, none of the generated sentences appear in the training set of COCO (100\% on \%Novel). 

CLIPScore~\cite{hessel2021clipscore} is a reference-free method for evaluating relatedness between an image and its caption, using CLIP's alignment score. Evidently, our method is much better in this metric than the supervised method (87\% vs. 77\%). As an alternative to exact correspondence with human reference, we use CLIPScoreRef to measure the semantic distance from the references. Although supervised methods outperform our method in this score (similarity in the vocabulary and the sentence style still provide an advantage), the gap is narrower than in other supervised metrics. 

\begin{figure*}[t]
	\centering
    \includegraphics[width=1\linewidth]{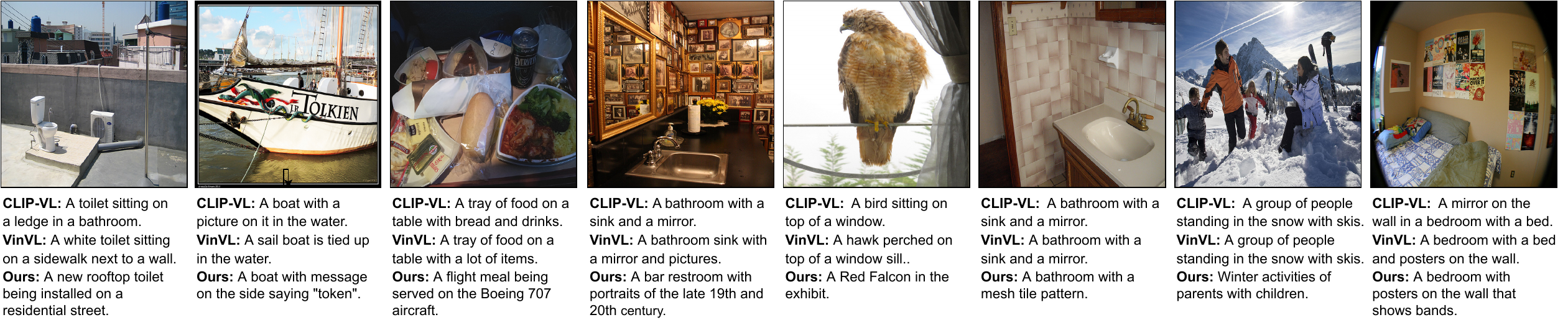}\vspace{-5pt}
    \caption{Examples of our zero-shot image captioning compared against 
    supervised captioning methods.}\vspace{-15pt}
    \label{fig:examples}
\end{figure*}

\begin{figure}[t]
	\centering
    \includegraphics[width=1\linewidth]{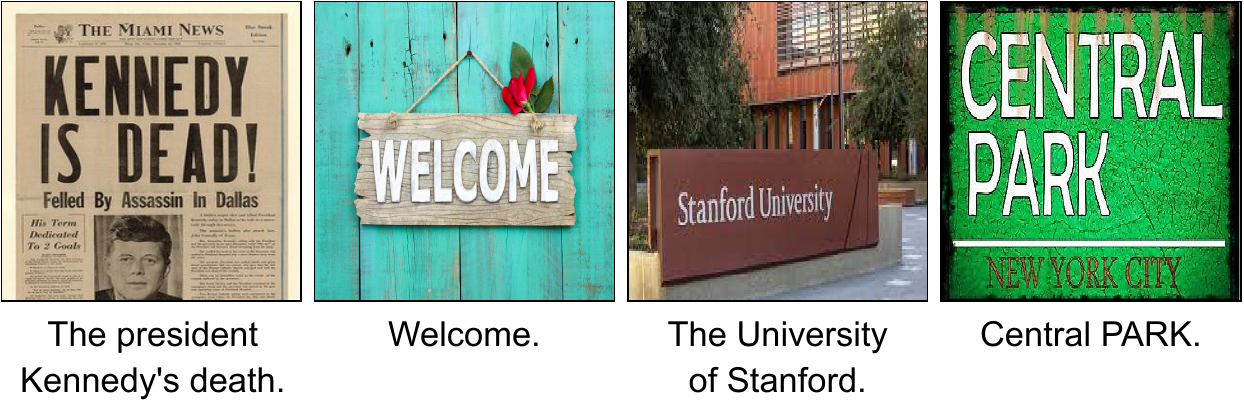}\vspace{-5pt}
    \caption{Examples of OCR capabilities.}\vspace{-8pt}
    \label{fig:ocr}
\end{figure}

\begin{figure}[t]
	\centering
    \includegraphics[width=1\linewidth]{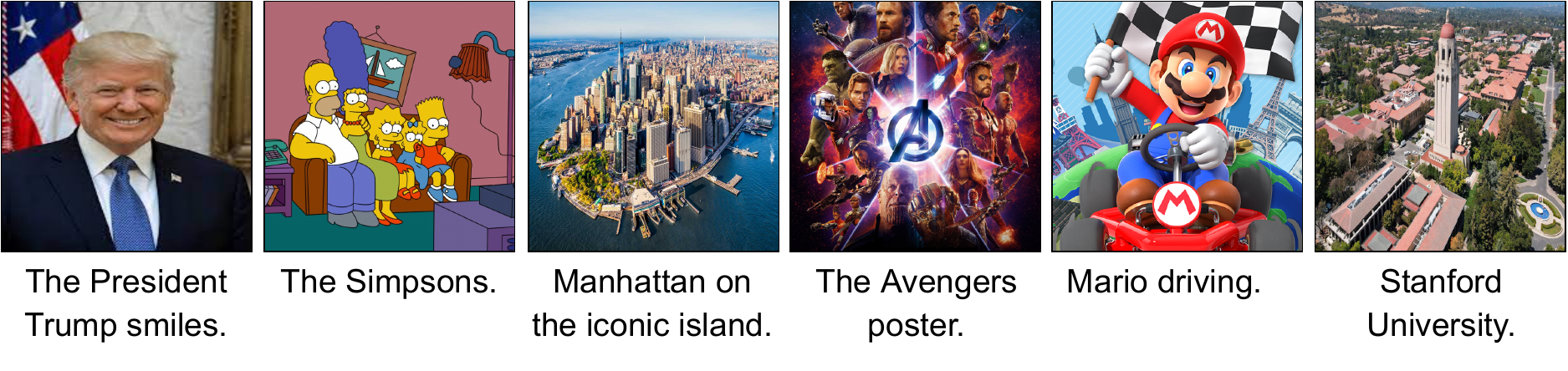}\vspace{-5pt}
    \caption{Examples of real-world knowledge.}\vspace{-16pt}
    \label{fig:real}
\end{figure}

\noindent{\bf Qualitative Analysis\quad} \cref{fig:examples} compares our zero-shot approach with other baselines, demonstrating that our method can generate human-like captions, i.e., textually richer, better at image reasoning, and more effective at grounding objects. We discuss each image from left to right. First, as opposed to CLIP-VL, which assumes a toilet is in the bathroom, and VinVL, which disregards the background buildings and presumes it is on a sidewalk, our method determines it is on a rooftop. Next, our method attempts to generate the written text on a boat's side. The following image describes a flight meal as a regular tray of food with the baselines, whereas our method describes it as a flight meal. We accurately describe the next image as a bar restroom with portraits and not a bathroom. Our method and VinVL specify specific birds in the following photo (red falcons and hawks are hard to tell apart). Next, the baselines repeat the same sentence, while our method mentions an interesting mesh tile pattern. In the next photo, our method identifies a family rather than a general group. Last, our method accurately describes a room's interior, such as a bedroom with posters, and deduce that the posters depict bands. Note that the baselines' captions are generally of the same pattern, while our method generates novel sentences. Also, note that the images are taken from COCO dataset, which was used to fine-tune CLIP-VL and VinVL.

\noindent{\bf OCR\quad} The ability of CLIP to classify text within an image from a closed set of possible prompts is impressive~\cite{radford2021learning}.  We show in \cref{fig:ocr} that these capabilities can be exploited in a generative manner. To accomplish this, we change the prefix prompt we use in our method from ``Image of a'' to ``Image of text that says.'' Results include impressive understanding. e.g., ``The president Kennedy's death'' from an image of a paper declaring it or generating ``The University of Stanford'' from a sign depicting its name.

\noindent{\bf External knowledge\quad} The generated captions comprise a wealth of real-world knowledge on a variety of topics. In \cref{fig:real} we show samples of famous people (e.g., Trump), animated shows (e.g., Simpsons), cities (e.g., Manhattan), movies (e.g., Avengers), games (e.g., Mario driving), and places (e.g., Stanford). 

  \subsection{Visual-Semantic Arithmetic Study}
\label{sec:res-arit}

We demonstrate how our method can generate text for subtraction to explain semantic directions. Next, we demonstrate that the summation operator allows guidance of the generated text through visual cues. One can then apply the above insights to solve visual analogy puzzles. 

\begin{figure*}[t]
	\centering
    \includegraphics[width=1\linewidth]{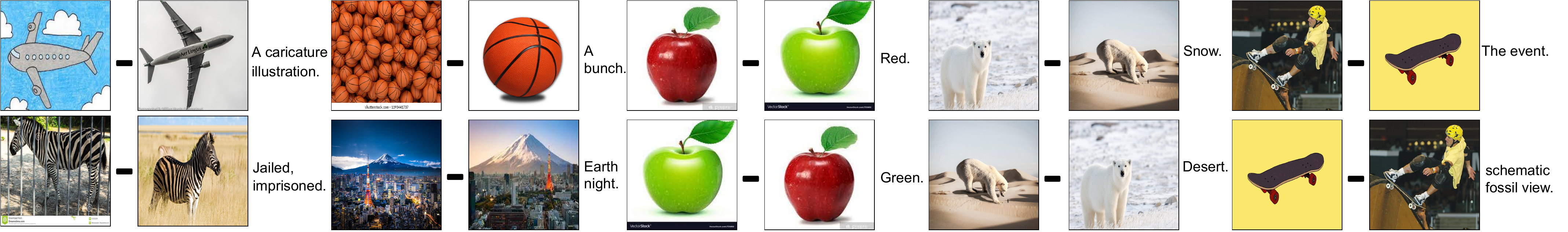}\vspace{-5pt}
    \caption{Examples of vector directions derived by subtracting representations from CLIP's embedding space. By generating text for the given direction, the concept is revealed.}\vspace{-5pt}
    \label{fig:subtraction}\vspace{-10pt}
\end{figure*}

\noindent{\bf Subtraction\quad} Subtracting vectors intuitively represents a direction between the vectors. In  \cref{fig:subtraction} we demonstrate our method's ability to express relations through several examples. ``A caricature illustration'' is the result of subtracting a real photo of an airplane from a caricature. To put it another way, adding the concept ``A caricature illustration'' to the right image of a real plane will match the image on the left of a caricature plane. Concepts of quantity and color can also be seen, for example, a comparison of a green apple versus a red apple yields `Red,' and vice-versa, subtracting one basketball from many basketballs results in ``a bunch.'' Furthermore, we find directions related to a geographical area, e.g., `Snow,' and 'Desert.` Further, a concept directly tied to day and night, and a concept of prison (i.e., `Jailed').  It should be noted that the operator is not symmetric, and cannot always be derived textually. For instance, on the right images, the concept direction from a skateboard to a skateboard tournament can be generated as ``The event.'' However, 
the direction from a skateboard tournament to a skateboard generated ``schematic fossil view,'' which is irrelevant. 

\begin{figure}[t]
	\centering
    \includegraphics[width=1\linewidth]{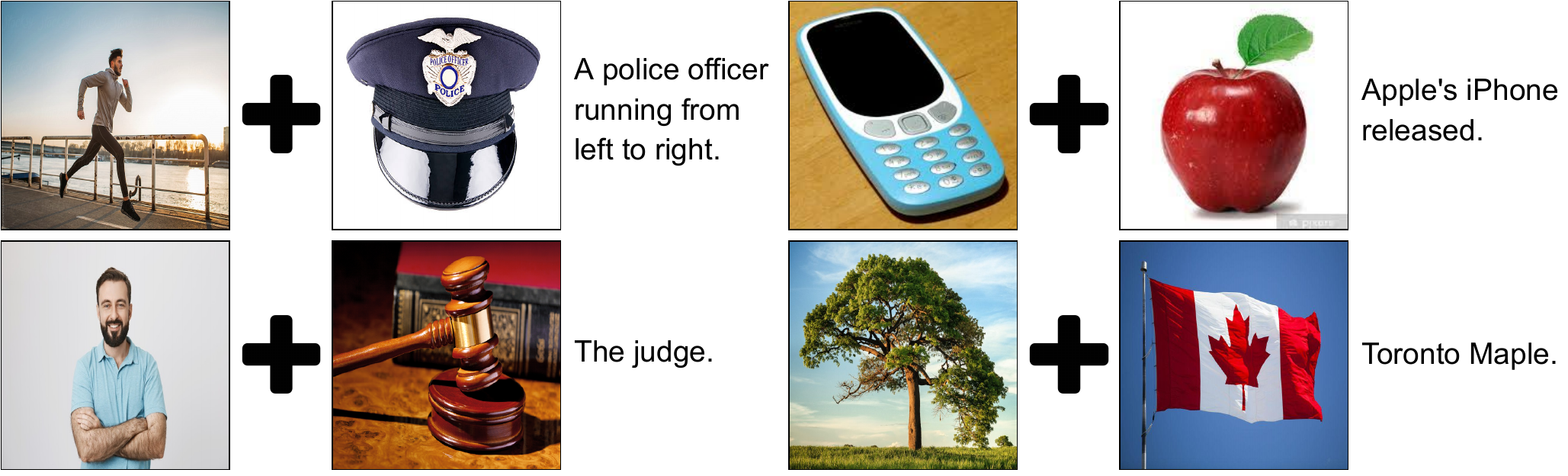}\vspace{-5pt}
    \caption{Examples of caption guidance with an image through the addition operator. }\vspace{-5pt}
    \label{fig:sum}\vspace{-10pt}
\end{figure}

\noindent{\bf Summation\quad} Through the addition operation, the generated text can be guided through visual semantics. In \cref{fig:sum} we show examples of guidance. On the left side, with the addition of a police officer's hat, the caption describes a man running as ``A police officer...,'' if we add a hammer to a man, we get ``The judge.'' On the right side, we show that a concept can be abstract. For example, the Apple company can be represented by an apple. Thus, adding an apple to a phone, results in the text ``Apple's iPhone released.'' Additionally, a country's concept can be represented visually with flags. If Canada's flag is added to a tree, ``Toronto Maple'' results. 


\noindent{\bf Guidance with Visual-Relations\quad}
In the field of natural language processing, semantic relations have long been studied~\cite{mikolov2013efficient}.  Previous efforts studied visual relations with expensive annotated language-priors~\cite{lu2016visual}. With the introduction of CLIP, richer visual concepts from large-scale data became available~\cite{goh2021multimodal}. Through visual arithmetic, we are able to exploit this richer embedding space. 

\begin{figure*}[t]
	\centering
    \includegraphics[width=1\linewidth]{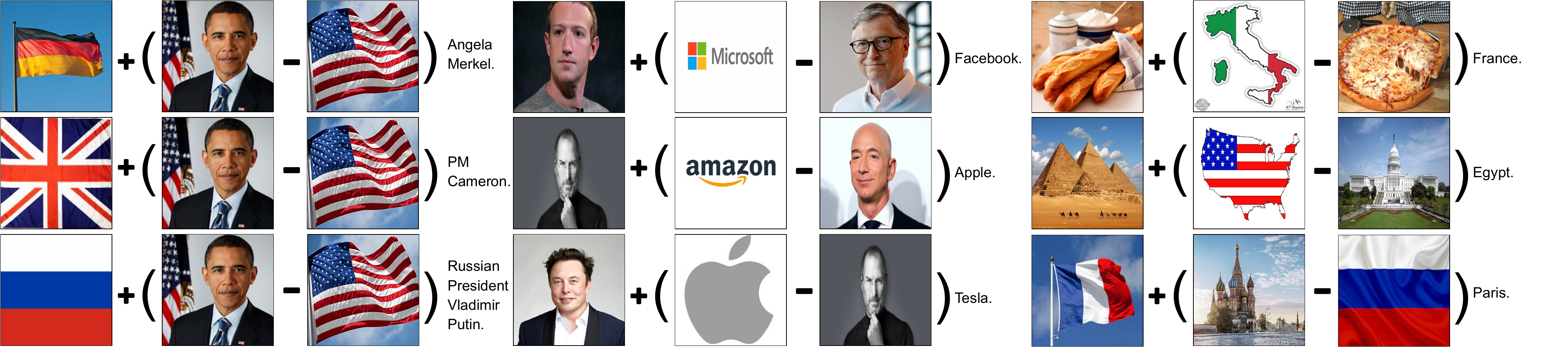}\vspace{-5pt}
    \caption{Image arithmetic with both summation and subtraction. For example, on the left side, by removing the American flag from Obama, a leadership direction results. The presidents of different countries are generated when the derived vector is added to their flags.}\vspace{-5pt}
    \label{fig:relations}
\end{figure*}

In \cref{fig:relations}, we show our proposed strategy. Using subtraction, we first determine the direction. For example, the concept of leadership is represented by an image of Obama minus the American flag. With this direction in hand, we can now manipulate the case of other nations. By adding the direction to the German flag, we obtain ``Angela Merkel.'' 
A different example is to examine the concept direction of CEO-to-company. With different images (e.g., Bill Gates and Microsoft, Jeff Bezos and Amazon), the direction can be summed to Mark Zuckerberg and Steve Jobs generating `Facebook' and `Apple,' respectively.  On the right side, we study various interactions with country-related representations. We guide the image of a baguette to generate `France' by taking photos of pizza and Italy and deriving the country-to-food direction. 

\noindent{\bf The Visual Relations benchmark\quad} To further study the relation capabilities of our technique quantitatively, we introduce a new benchmark of visual relations, VR for short. This benchmark comprises 320 relations of the following templates: buildings$\rightarrow$country, countries$\rightarrow$capital, foods$\rightarrow$country, leaders$\rightarrow$country, and CEO$\rightarrow$company. These were chosen because they are roughly many to one, i.e., a country has many buildings, but a building only relates to one country.  The benchmark is designed to measure both the ability to model relations visually and to apply  real-world knowledge to perform the task. 

We constructed the benchmark through the following steps: (i) we created semantic directions by subtracting visual pairs and (ii) we then used each direction and added it to a visual element in another pair to create its corresponding text companion. As an example, we used images of ('japan,' 'sushi') to convey the direction of food$\rightarrow$country, and then we added this direction to an image of a pizza and examined the appearance of Italy in the generated text. 

We focused on single-word answers. 
The three evaluation metrics we find relevant to this setting are (1) BLEU-1, which measures unigram precision; (2) Recall@5, which indicates a word's appearance within the first five words generated; and (3) CLIP-score, which indicates semantic relatedness. To calculate the CLIP-score, we first add ``Image of'' as a prefix to the ground truth. Using CLIP's textual encoder, we then use a cosine distance. More details are provided in the supplementary material. 

In \cref{tab:relationresults},  we show performance for each relation. While this task is challenging, our approach resulted in a significant success rate of 30\% at R@5 in most relations. Note that, since the benchmark lacks multiple references, it is still limited, e.g., we mark a miss if the generated word is `US,' while the ground truth is `USA'. Observing the returned answers reveals that some  mistakes are understandable, e.g., answering Sydney instead of Canberra or the Sinai province instead of the country Egypt. However, other cases return truncated sentences, e.g., returning `flag'  instead of a country name or returning general concepts such as ``flickr image''. See supplementary for a discussion.  When employing the softer CLIP-Score metric, which is based on a semantic distance, a correlation of 70\% is observed. 

We compared our results with ClipCap~\cite{clip-prefix} that encodes the image with CLIP's image encoder and uses it as an initial token for GPT-2. The method is fine-tuned based on COCO dataset. As can be seen, this method fails to retrieve the correct response, despite employing the same large-scale models as we do and performing arithmetic in the same CLIP embedding space. CLIP-VL~\cite{shen2021much} and the supervised captioning methods cannot be tested on this benchmark since it uses spatial grid features as embedding. 

\begin{table*}[t!]
{
\centering
    \resizebox{\linewidth}{!}{%
\begin{tabular}{lcccccccccccccccc} 
\toprule

& \multicolumn{3}{c}{\textbf{Building $\rightarrow$ Country}} & \multicolumn{3}{c}{\textbf{Country $\rightarrow$ Capital}} & \multicolumn{3}{c}{\textbf{CEO $\rightarrow$ Company}} & \multicolumn{3}{c}{\textbf{Food $\rightarrow$ Country}} & \multicolumn{3}{c}{\textbf{Leader $\rightarrow$ Country}} \\ 

\cmidrule(lr){2-4} 	\cmidrule(lr){5-7} \cmidrule(lr){8-10} \cmidrule(lr){11-13}
\cmidrule(lr){14-16} 

Method & B@1 & R@5 & C-s & B@1 & R@5 & C-s & B@1 & R@5 & C-s & B@1 & R@5 & C-s & B@1 & R@5 & C-s   \\ 

\midrule

ClipCap \cite{clip-prefix}  & 0.003 &  0.035  & 0.24 & 0.0 &  0.0  & 0.22 & 0.004 &  0.05  & 0.18 & 0.0 &  0.0  & 0.24 & 0.008 &  0.24  & 0.26 \\

Ours  & \textbf{0.1} &  \textbf{0.32}  & \textbf{0.7} & \textbf{0.14} &  \textbf{0.32}  & \textbf{0.68} & \textbf{0.1} &  \textbf{0.3}  & \textbf{0.64} & \textbf{0.03} &  \textbf{0.33}  & \textbf{0.66} & \textbf{0.1} &  \textbf{0.28}  & \textbf{0.68} \\ 
	
\bottomrule
\end{tabular}}
\vspace{-6pt}
\caption{Comparison of our method and ClipCap baseline on our novel benchmark for visual relations. B@1 = BLEU-1, R@5 = Recall@5, C-s = CLIP-score.}
\vspace{-18pt}
\label{tab:relationresults}}
\end{table*}


\begin{figure}[t]
	\centering
    \includegraphics[width=1\linewidth]{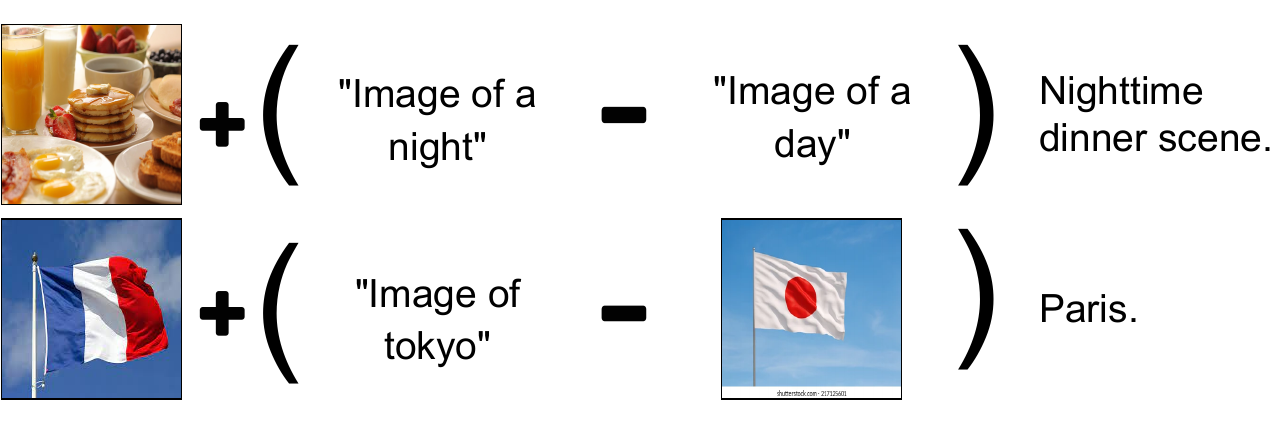}\vspace{-7pt}
    \caption{Our method is not only devoted to visual relations, but it also allows arithmetic between image and language.}\vspace{-15pt}
    \label{fig:mm}
\end{figure}

\noindent{\bf Multi-modal Arithmetic\quad}
Our method enables multi-modal reasoning, which involves manipulating images and text simultaneously in the same embedding space. Using CLIP's textual encoder, $E_{\text{Text}}$. In \cref{fig:mm}, we show that a day-to-night direction can be obtained with text inputs, i.e., ``image of a night,'' and ``image of a day.''  The direction steers an image of breakfast to ``Nighttime dinner.'' 

\section{Discussion and Limitations}
\label{sec:limit}
The zero-shot capabilities presented by CLIP~\cite{radford2021learning} pave a new path for computer vision. However, these are limited to multiclass classification. DALL-E~\cite{ramesh2021zero} presents an impressive ability to generate images that are very different from its training images in what is termed zero-shot generation ability. However, this ability is exactly the generative task DALL-E was trained to do, only in new domains.  No previous computer vision work, as far as we can ascertain, has presented a generative semantic zero-shot capability of the sort that is revolutionizing the NLP world with transformers, such as GPT-3~\cite{brown2020language}. Our work is the first to present a generative visual-semantic work.
 
While the ability to rely on pre-trained models such as GPT-2~\cite{radford2019language} and CLIP allows us to achieve such new capabilities, they also highlight the uneven playing field AI has become. GPT-2 is far inferior to GPT-3 and other recent LMs in which resources far beyond the reach of most research labs are invested. 

On a similar note, it is likely that combining zero-shot with supervised training would lead to a method that outperforms the baselines in all captioning metrics. 
However, the amount of resources currently used to train supervised captioning methods is becoming a deterring factor from pursuing this direction. For instance, UNITER uses 3645 hours of a V100 GPU~\cite{chen2020uniter}.

The use of an LM and an image-language matching model trained on large corpora of collected data inevitably leads to biases. For example, the models we employ are clearly oriented towards Western knowledge and can recognize people, places, objects and concepts that are popular in Western media, while being much less knowledgeable about other cultures. For example, 
our model fails to form relations with the president of China, Xi Jinping.

\section{Conclusions}

The marriage between a language model and a visual-semantic matching model is a powerful union, with the potential to provide zero-shot captioning that brings together real-world variability in text, recognition abilities that are unrestricted by categories, and real-world knowledge that is embedded in the models through web-scale datasets.

We propose a zero-shot method for combining the two models, which does not involve optimizing over the weights of the models. Instead, we modify, for all layers and attention heads, the key-value pairs of the tokens generated by the language model up to each inference step. 

As a captioning model, our method produces results that are less restrictive than those provided by the human annotators on the datasets used by supervised captioning methods. While this lowers the word-to-word metrics, the captions generated seem to be a good match to the image at the semantic level and exhibit real-world information. Moreover, the flexibility of using an embedding-space zero-shot method enables us to perform visual-semantic arithmetic. 

We show how we can describe in words the difference between two images and how we can combine concepts from multiple images. Both are novel high-level recognition tasks. Combining these two capabilities, a powerful image analogy machine is obtained, which answers, by providing a text string, questions of the form  ``A is to B as C is to X'' ($X\sim C+B-A$), in which A, B, and C can each be either textual or visual. 


\section*{Acknowledgments}
This project has received funding from the European Research Council (ERC) under the European Unions Horizon 2020 research, innovation programme (grant ERC CoG 725974). The contribution of the first author is part of a PhD thesis at Tel Aviv University.

{\small
\bibliographystyle{ieee_fullname}
\bibliography{main_arxiv}
}

\appendix
\section{Supplementary Material:  ZeroCap: Zero-Shot Image-to-Text Generation for Visual-Semantic Arithmetic}

\maketitle
\thispagestyle{empty}

This supplementary material describes our experimental setup (see \cref{sec:exp}), provides additional ablation study (see \cref{sec:ablations}), provides additional qualitative results (see \cref{sec:qual}), explores the limitations of our approach (see \cref{sec:limit_app}), and discusses visual relation benchmark failure cases (see \cref{sec:relation}).

\section{Experimental Setup}
\label{sec:exp}

As part of our experiments, we used COCO's validation set (Karpathy splits) for both qualitative and quantitative evaluations. We report the beam with the lowest CLIP loss score among the five beams. Our model has several hyperparameters: (i) $\lambda$ (see Eq.~(2)), which was set to 0.2; (ii) $\tau_c$ (see Eq.~(3)), which was set to 0.01; (iii) $\alpha$ (see Eq.~(5)), which was set to 0.3;  (iv) We decreased the likelihood of repeated tokens by a factor of two in order to mitigate repetitions. Based on a human assessment, these parameters produced concise, fluent, and image-related captions. 
We use the PyTorch framework~\cite{NEURIPS2019_9015}.

 \noindent{\bf Pre-trained models:\quad}As part of our approach, we use two large-scale pre-trained models: (i) GPT-2, using HuggingFace's gpt2-medium implementation\footnote{\url{https://huggingface.co/transformers/model_doc/gpt2.html}}, with 24 attention models and 345M trainable parameters. This model was trained on an 8M web-page dataset with a causal language modeling (CLM) objective; (ii) CLIP, trained on 400M (images, text) crawled from the web. We use the OpenAI implementation\footnote{\url{https://github.com/openai/CLIP}}. We employed a version of CLIP with a vision transformer image encoding architecture that is equivalent to ViT-B/32~\cite{dosovitskiy2020vit}.

\begin{table}[t!]
\centering
\begin{tabular}{lcc}
\toprule
{Method}& CLIP-S
& Perplexity
\\ \midrule
A1 & \textbf{0.98} & 8.61
\\
A2 & 0.91 & 6.04
\\ \midrule
Ours & 0.87 & \textbf{5.50}
\\
\bottomrule
\end{tabular}
\caption{Comparison of our method with and without optimization. We show two variants: (A1) selecting tokens one by one to maximize the CLIP score, and (A2) doing so on a score that combines CLIP score with an LM-score.}
\label{tab:optimization_table}
\end{table}

\noindent{\bf Prompt engineering:\quad}  Our method begins with an initial prompt. In the majority of our experiments, we used ``Image of a''. We determine the caption from the words generated after the initial prompt. We did observe that the prompt affected output results, e.g., ``Image of text that says,'' is much better if the caption is intended for OCR. 

\section{Ablation Study}
\label{sec:ablations}
\begin{figure*}[t]
	\centering
    \includegraphics[width=0.95\linewidth]{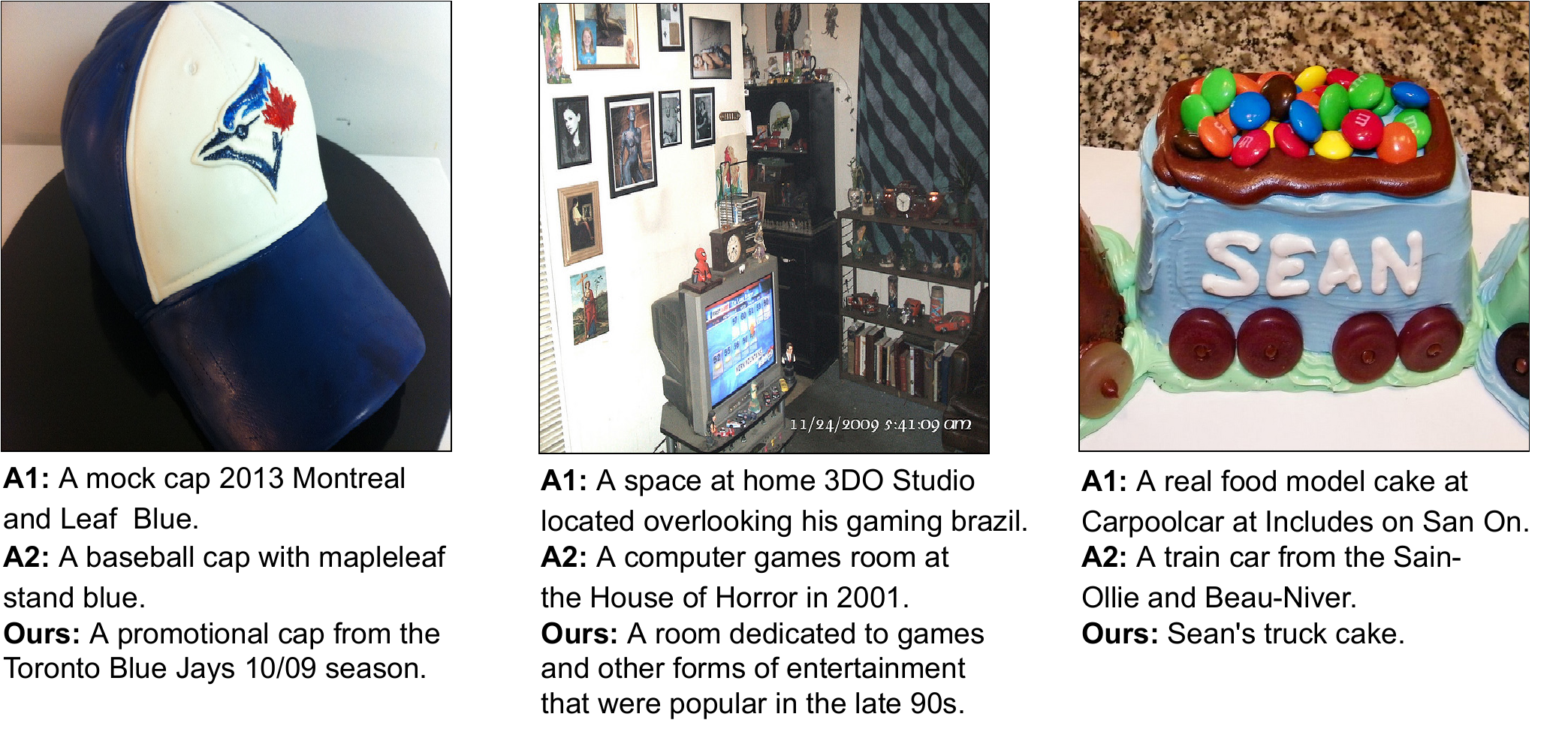}
    	\caption{Illustration of methods that employ CLIP directly without optimization to the LM. We show two variants: (A1) selecting tokens one by one to maximize the CLIP score, and (A2) doing so on a score that combines CLIP score with an LM-score.}
        \label{fig:ablation_teaser}
\end{figure*}

\noindent{\bf Effect of CLIP-based optimization:\quad} A further ablation was performed, in which CLIP's score is used directly to optimize the LM. In \cref{fig:ablation_teaser}, we show two variants: (A1) selecting tokens one by one to maximize the CLIP score, and (A2) doing so on a score that combines CLIP score with an LM-score. Evidently, the captions are not competitive with our method. We also assessed the differences in language fluency (perplexity measured with GPT Neo) and image correspondence (measured with CLIP Score). Despite a higher CLIP score (\cref{tab:optimization_table}), our method has improved language fluency. It is worth noting that higher CLIP doesn't necessarily translate to better wording.

A human study further supports this, conducted to determine which method is perceived as the best one. The study included 50 images randomly selected from COCO and 40 annotators. Our caption was selected 70.5\% , (A1)  8.9\%, and (A2) 20.6\%.

\noindent{\bf Effect of regularizer coefficient:\quad}As shown below, an increase in the regularizer coefficient results in a decrease in the perplexity score measured with GPT Neo (\ie, language fluency improves) while it decreases the clip similarity. We find  $\lambda = 0.2$ to be a good trade-off point.

\noindent\begin{minipage}{.5\linewidth}
  \centering
  \includegraphics[width=.9\linewidth]{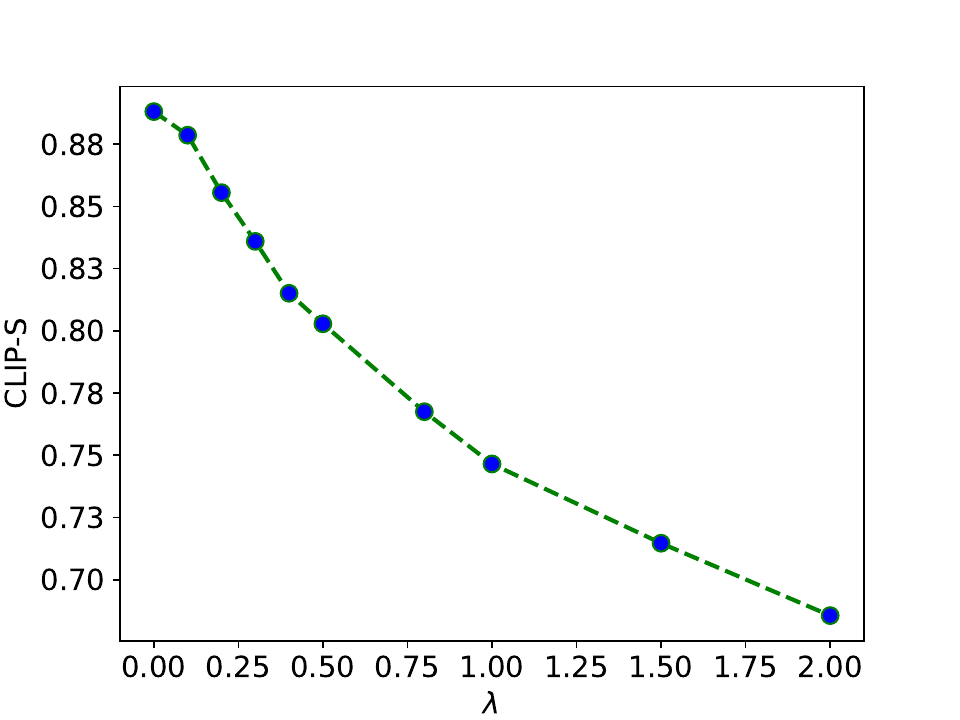}
  \label{fig:test1}
\end{minipage}%
\begin{minipage}{.5\linewidth}
  \centering
  \includegraphics[width=.9\linewidth]{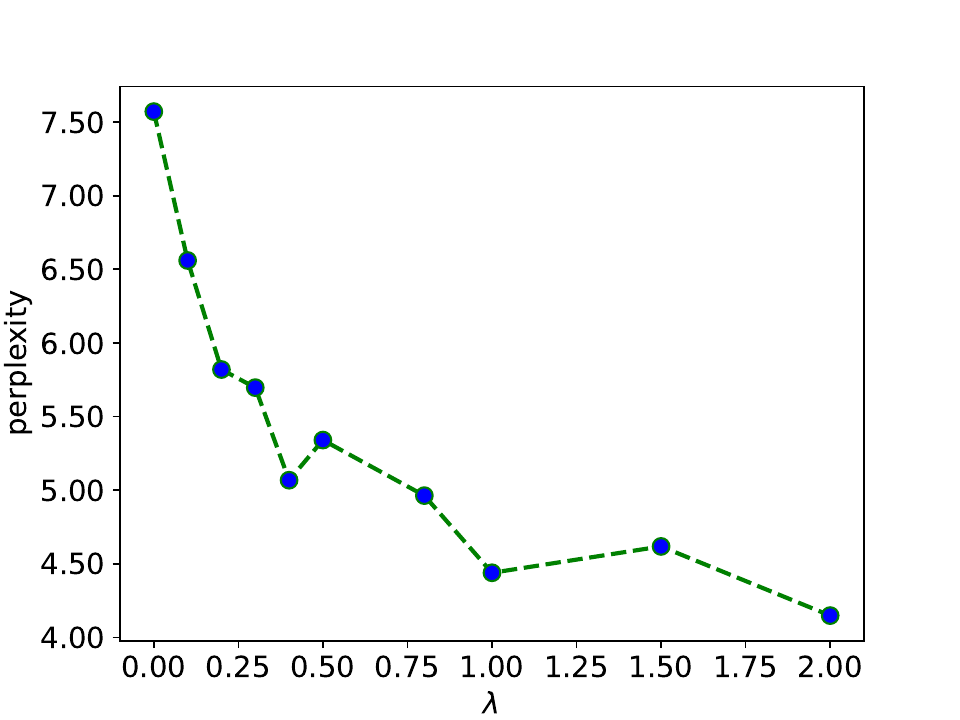}
  \label{fig:test2}
\end{minipage}

\noindent{\bf Human evaluation:\quad}We conducted an additional human study on 50 images. We picked the images from the web (\eg, video-game screenshot, real-world knowledge; specifically, the subreddit `i took a picture').
We asked the annotators to score between 1 to 5 two properties: human-like and visual grounding. We compared against a supervised method ClipCap. On human-like, our approach got 3.79 \vs 3.17 of ClipCap. On image grounding, our method got 3.98 \vs 3.21 of ClipCap.

\section{Additional Qualitative Results}
\label{sec:qual}

 \noindent{\bf Image Captioning:\quad} In \cref{fig:cap_examples} (shown at the end of the document due to size), we present our results on 200 randomly-selected images along with  baselines. For baselines, we use  ClipCap~\cite{clip-prefix}, CLIP-VL~\cite{shen2021much}, and VinVL~\cite{zhang2021vinvl}. Our method generates original captions that are completely different both in vocabulary and pattern from the baselines' captions.

\section{Limitations} 
\label{sec:limit_app}
We detail both the caption quality issues and the biases resulting from the noisy web-scale data used to train CLIP and GPT-2 in the following sections.

\begin{figure*}[t]
	\centering
    \includegraphics[width=0.95\linewidth]{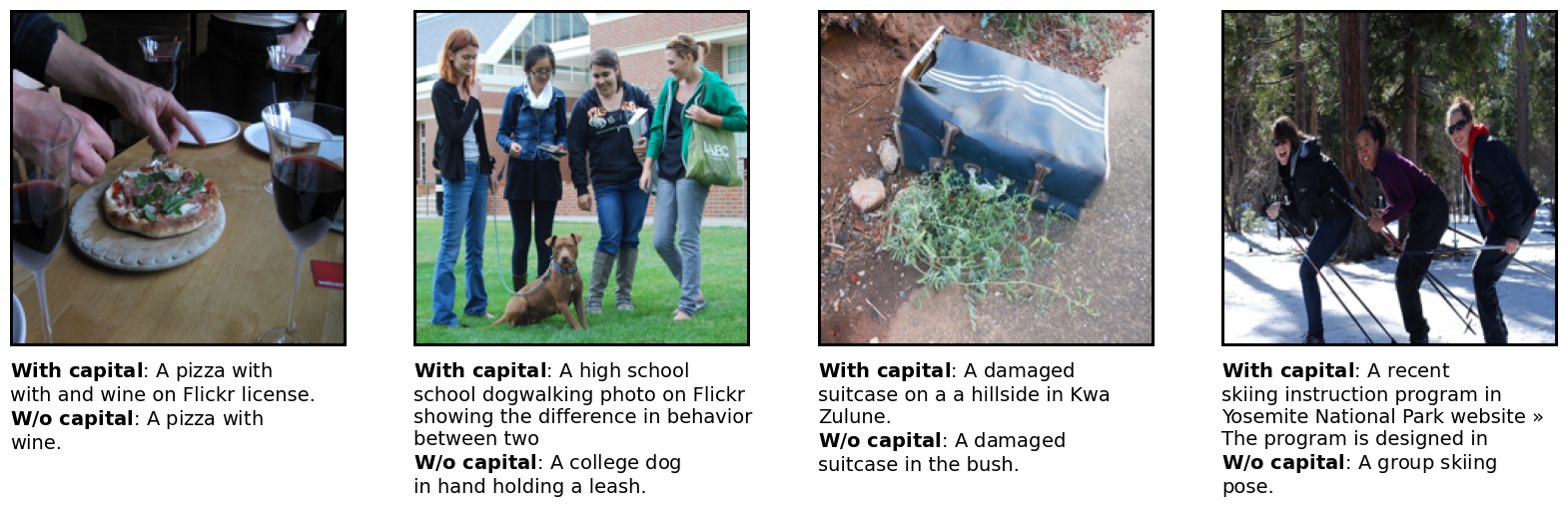}
    	\caption{The effect of our entity-control mechanism. With the mechanism (With Capital) and without the mechanism (W/O Capital).}
        \label{fig:long_shot}
\end{figure*}

\begin{figure*}[t]
	\centering
    \includegraphics[width=0.95\linewidth]{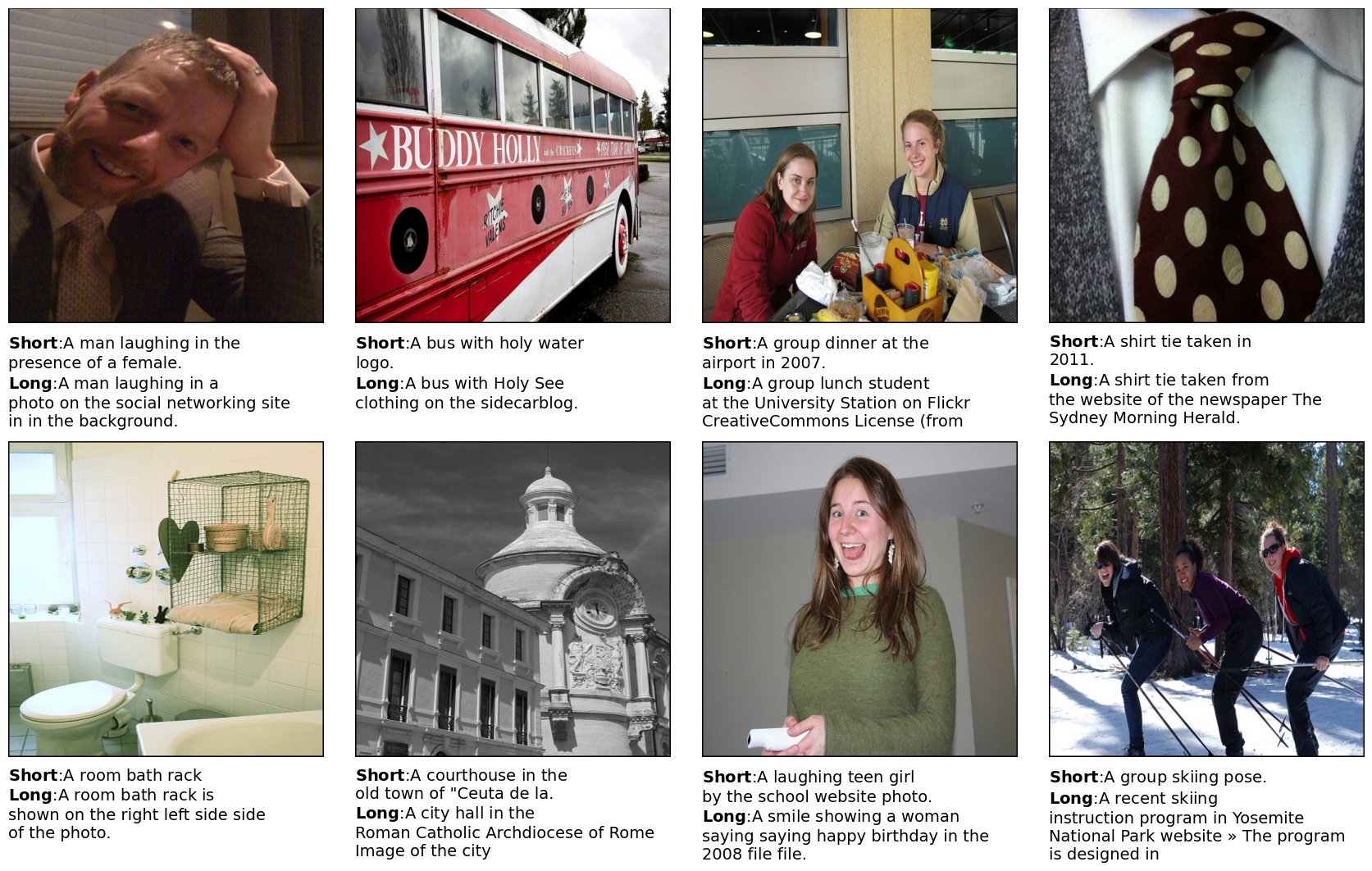}
    	\caption{The effect of our length-control mechanism. With the mechanism (Short) and without the mechanism (Long).}
        \label{fig:capital_examples}
\end{figure*}

\noindent{\bf Web-scale noise:\quad} The captions we generate are influenced by CLIP's training data. Due to its extraction from the web without special care, it contains noise. This leads to two undesirable outcomes: 1) Generating entities related to the data source (\eg, Flickr) or irrelevant entities (\eg, the name of the photographer). We solve this problem by adding a negative prior regularization to any capitalized subword. Consequently, a more generic caption will be created, but at the expense of world-knowledge capabilities. We show samples with and without the mechanism in \cref{fig:long_shot}; and 2) At times, the captions become irrelevant because they fail to remain focused.   This can be controlled using two hyperparameters. We multiply the probability of the end token by a factor of $f_e$, starting from time-step $t_e$. In our method we used $f_e = 1.04$, and $t_e=3$. In \cref{fig:capital_examples}, several random examples are shown, and the length control mechanism is ablated.

\begin{figure*}[t]
	\centering
    \includegraphics[width=0.95\linewidth]{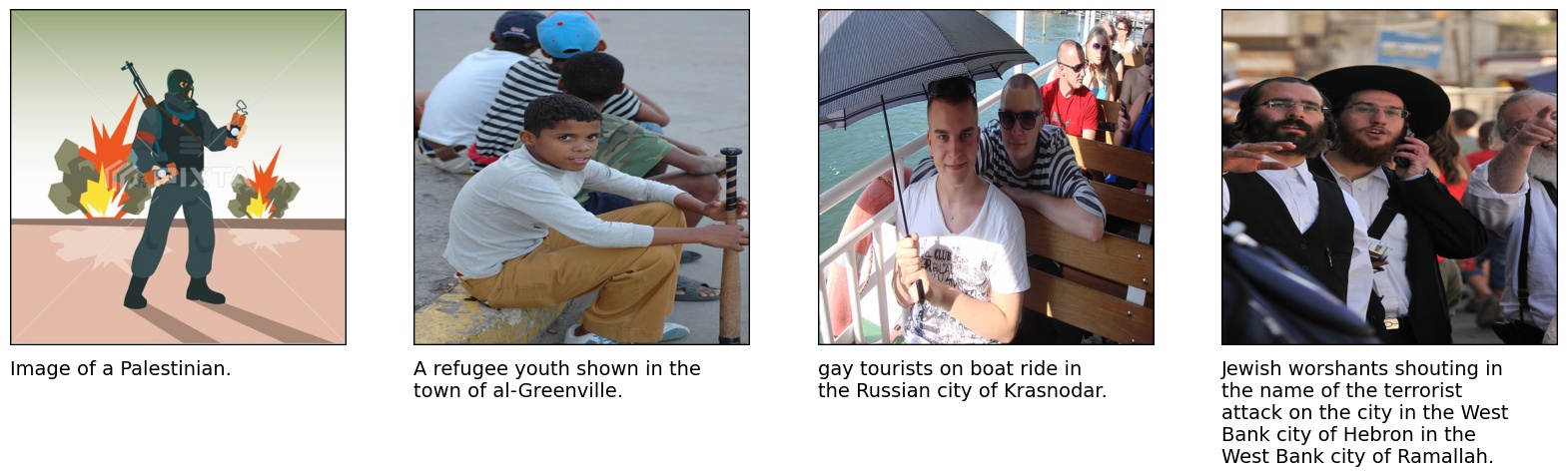}
    	\caption{Bias cases against distinct groups.}
        \label{fig:bias_examples}
\end{figure*}

\noindent{\bf Bias and Fairness:\quad} It is common for web-scale data to contain biased sources (\eg, news), resulting in an unintended bias against some ethnic groups. In \cref{fig:bias_examples},  an abstract illustration of a terrorist is described as Palestinian. Another example, racial characteristics are used to portray a child as an immigrant. Additionally, a caption implies homosexual orientation for an image of two men.

\section{Visual Relations Benchmark Study}
\label{sec:relation}

Our benchmark combines real-world knowledge with the ability to represent visual relationships. In \cref{fig:VR_examples}. we show at typical mistakes. Samples are referred to by their character counter: (a) Unpopular real-world knowledge. GPT-2 and CLIP training are based on web crawled data. Consequently, it may choose words based on popularity on the Internet. Sydney is a more popular city than Canberra worldwide (we validate this with Google Trends); (b) Synonyms. The relationship between the president and his or her country leads to "Canadian" rather than "Canada;" (c) Closely related. Rather than relating the pyramids to Egypt, this sample refers to Sinai, an area in Egypt; (d), (e),(f) Relation mistake. Subtracting Australia from Canberra conveys a relationship relevant to a university. It appears that adding the relationship to the UK led to `Berkeley.' A `Chinese university' is generated by adding it to China, and a `German university' is generated by adding it to Germany. This might be due to Canberra being known for its university. Since we use the same relation (pair subtraction) for multiple triplet of images, inferring the wrong relation can lead to many errors in the benchmark. 




\begin{figure*}[t]
	\centering
	\subfloat[]{{\includegraphics[width=0.45\textwidth]{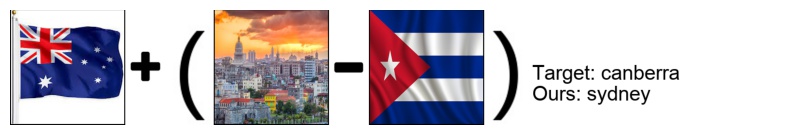} }}%
    \qquad
    \subfloat[]{{\includegraphics[width=0.45\textwidth]{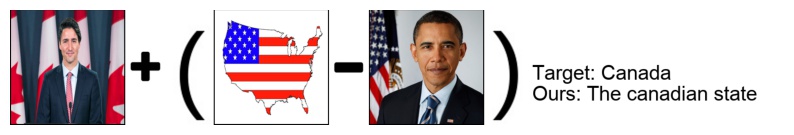} }}%
    \label{fig:VR_examples}
\end{figure*}

\begin{figure*}[h!]
	\centering
    \ContinuedFloat
    \subfloat[]{{\includegraphics[width=0.45\textwidth]{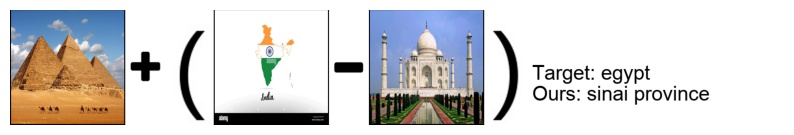} }}%
    \qquad
    \subfloat[]{{\includegraphics[width=0.45\textwidth]{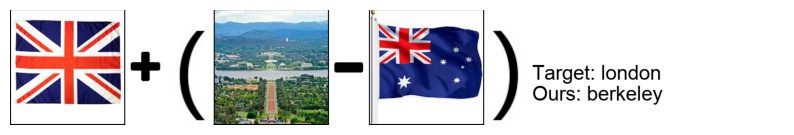} }}%
\end{figure*}

\begin{figure*}[h!]
	\centering
    \ContinuedFloat
    \subfloat[]{{\includegraphics[width=0.45\textwidth]{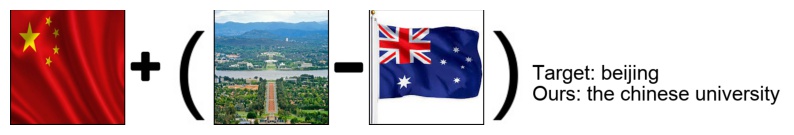} }}%
    \qquad
    \subfloat[]{{\includegraphics[width=0.45\textwidth]{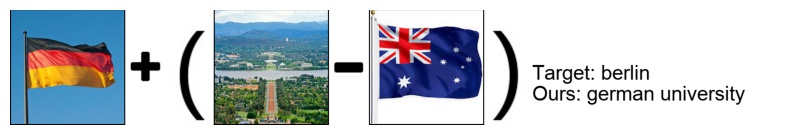} }}%
    \caption{Error analysis of the visual relations benchmark.} 
\end{figure*}


\begin{figure*}[t]
	\centering
	\caption{Generated captions by our method and by the baseline methods for images from the MS-COCO~\cite{ty2014coco} test-set. CP=ClipCap~\cite{clip-prefix}, CVL=CLIP-VL~\cite{shen2021much}, VVL=VinVL~\cite{zhang2021vinvl}.}
    \includegraphics[width=0.95\linewidth]{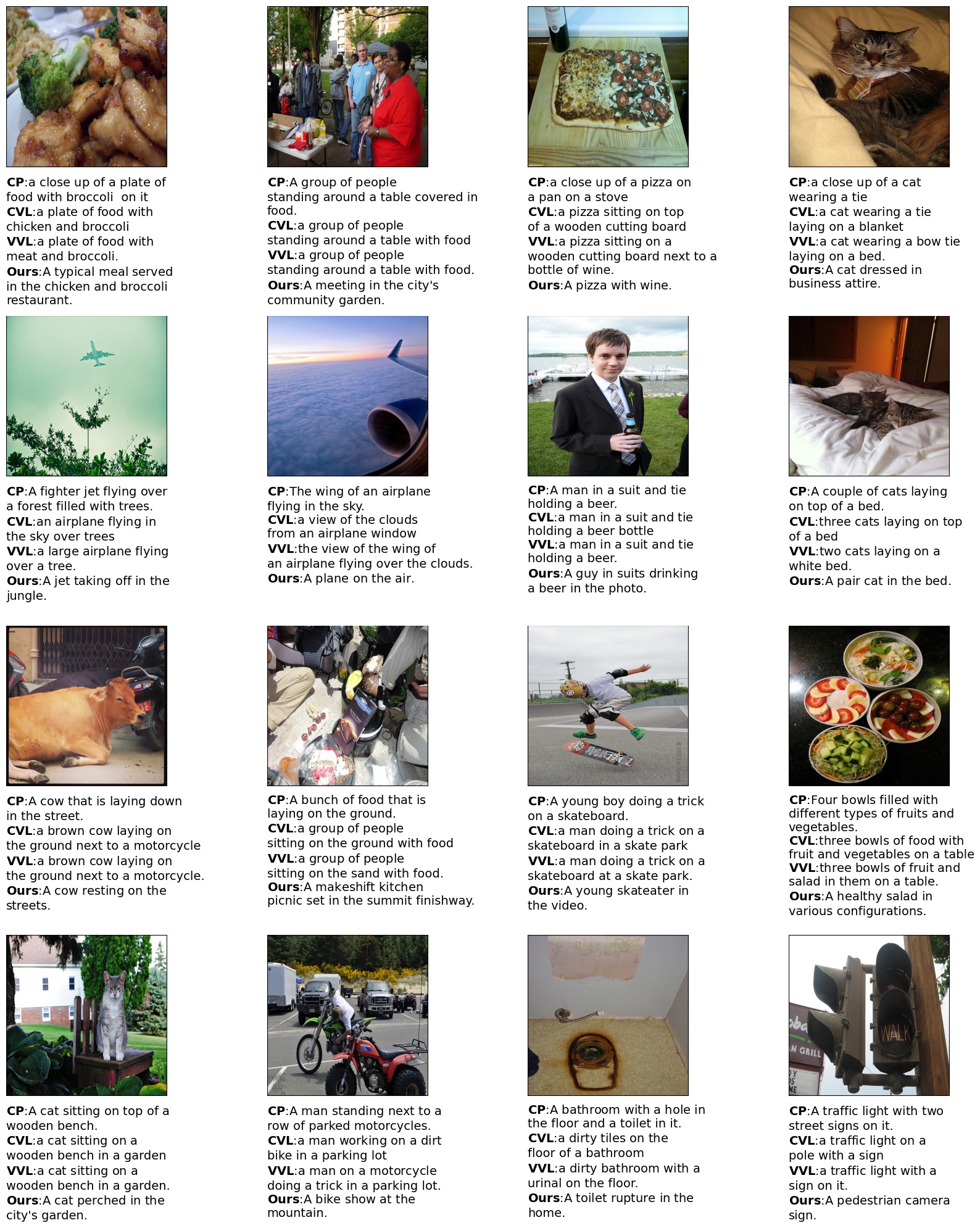}
        \label{fig:cap_examples}
\end{figure*}
\begin{figure*}[h!]
	\centering
    \ContinuedFloat
    \includegraphics[width=0.95\linewidth]{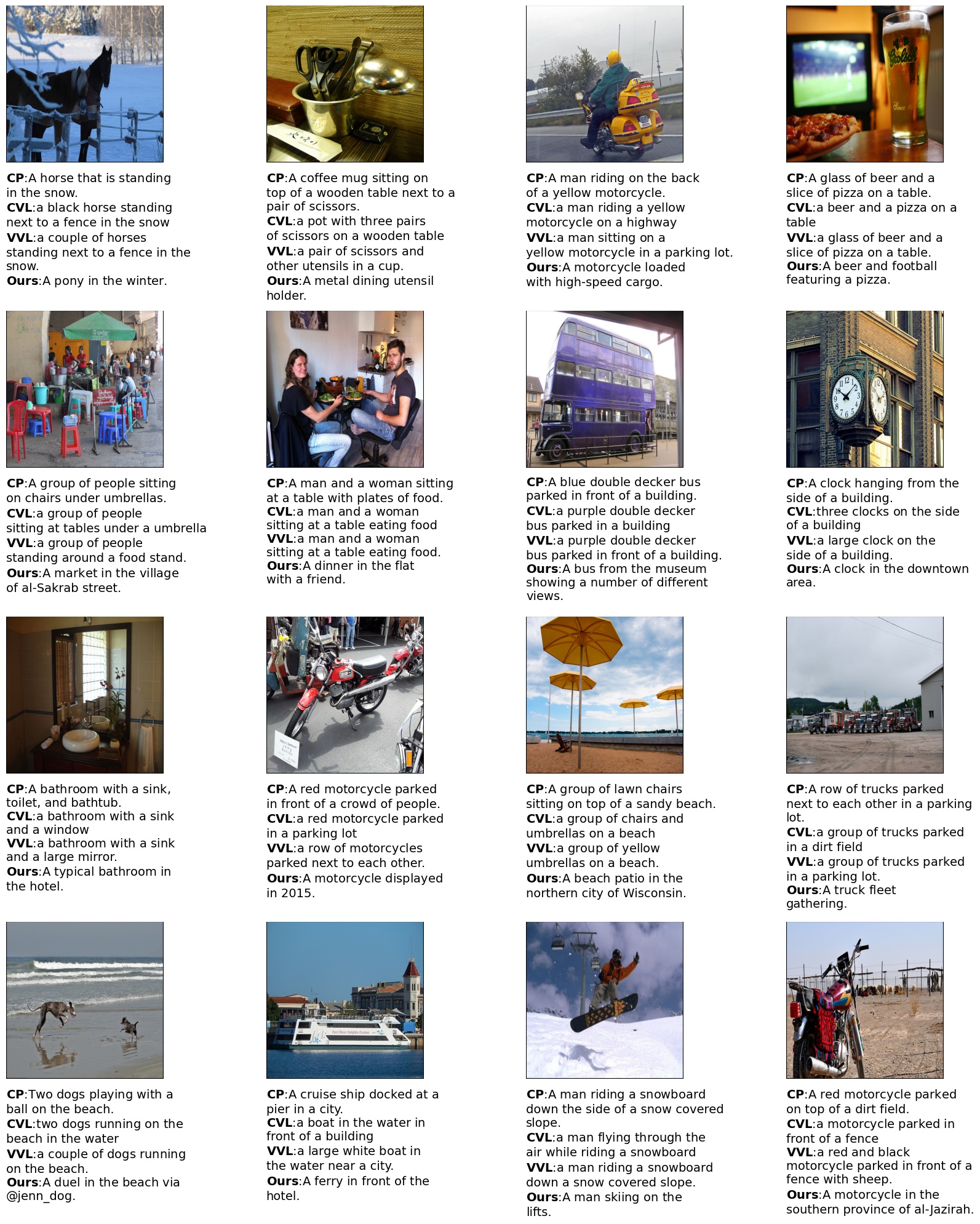}
\end{figure*}
\begin{figure*}[h!]
	\centering
    \ContinuedFloat
    \includegraphics[width=0.95\linewidth]{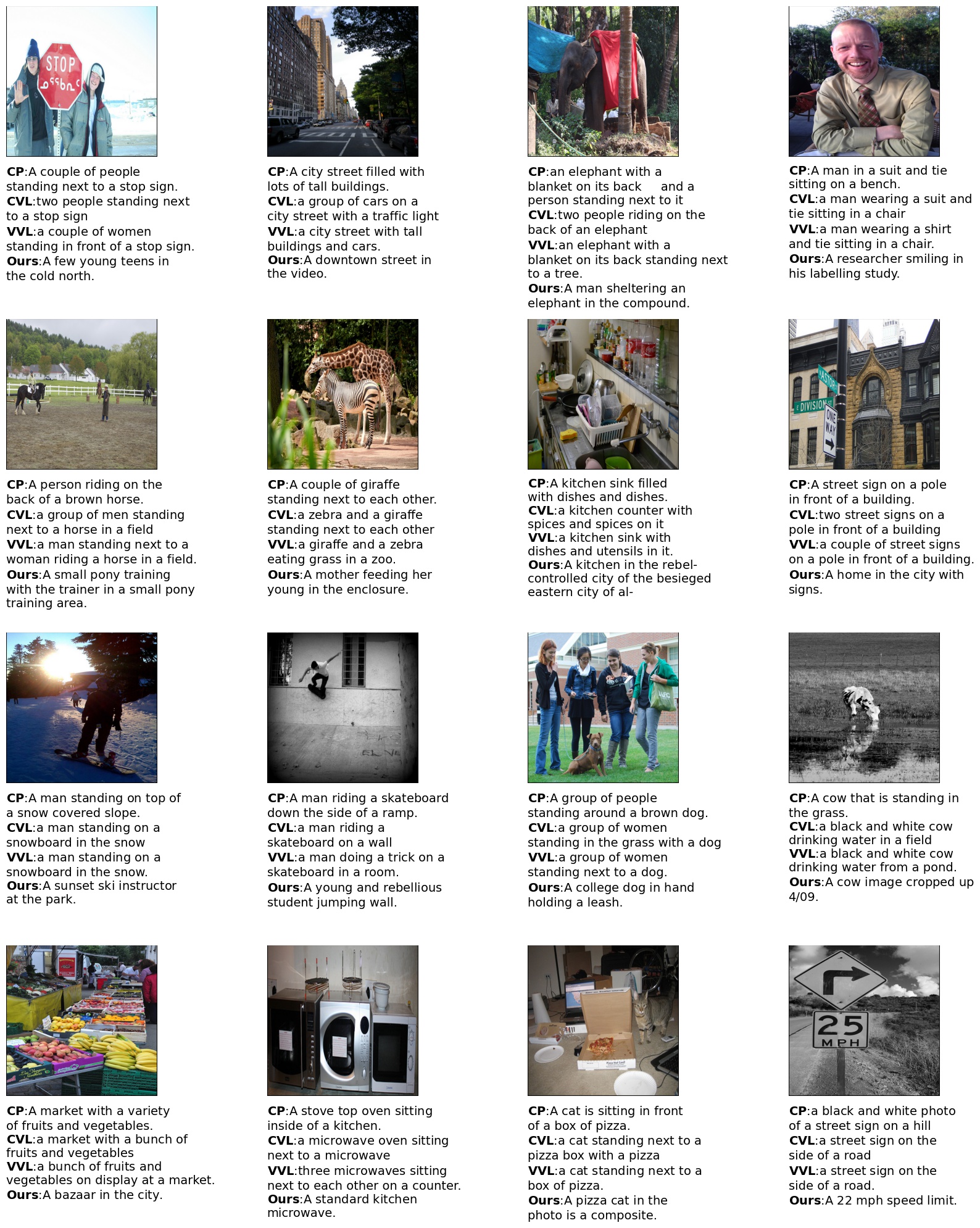}
\end{figure*}
\begin{figure*}[h!]
	\centering
    \ContinuedFloat
    \includegraphics[width=0.95\linewidth]{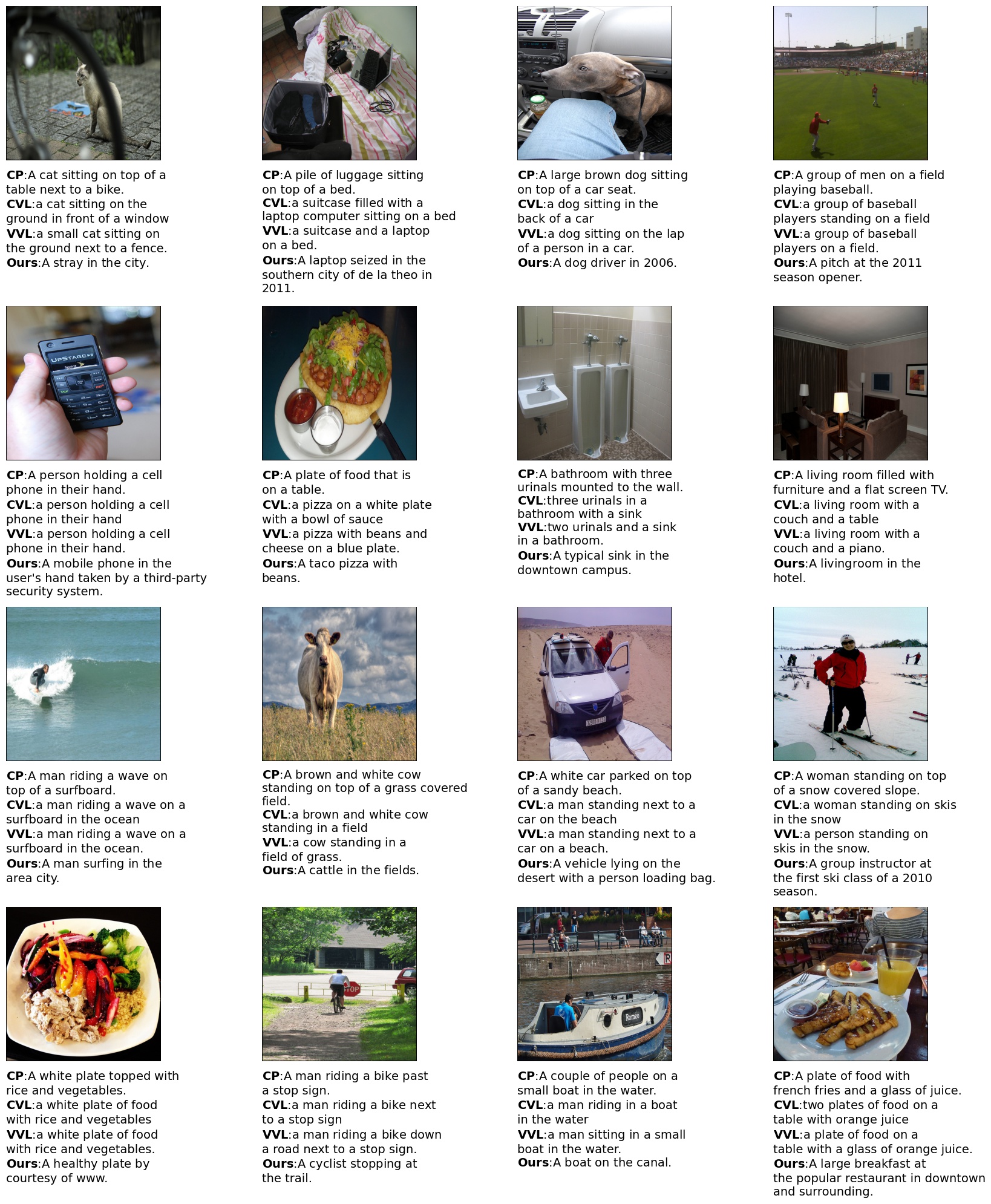}
\end{figure*}
\begin{figure*}[h!]
	\centering
    \ContinuedFloat
    \includegraphics[width=0.95\linewidth]{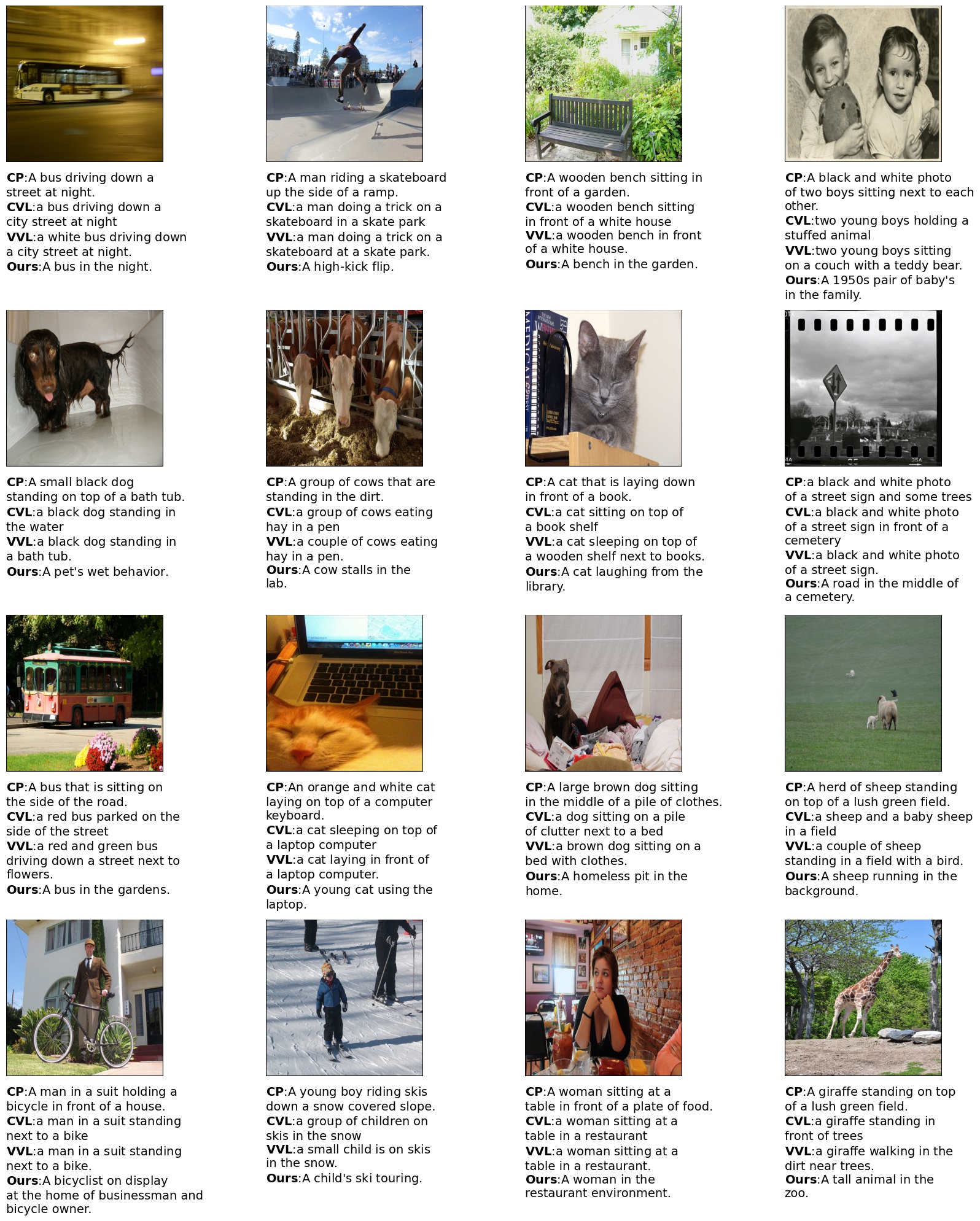}
\end{figure*}
\begin{figure*}[h!]
	\centering
    \ContinuedFloat
    \includegraphics[width=0.95\linewidth]{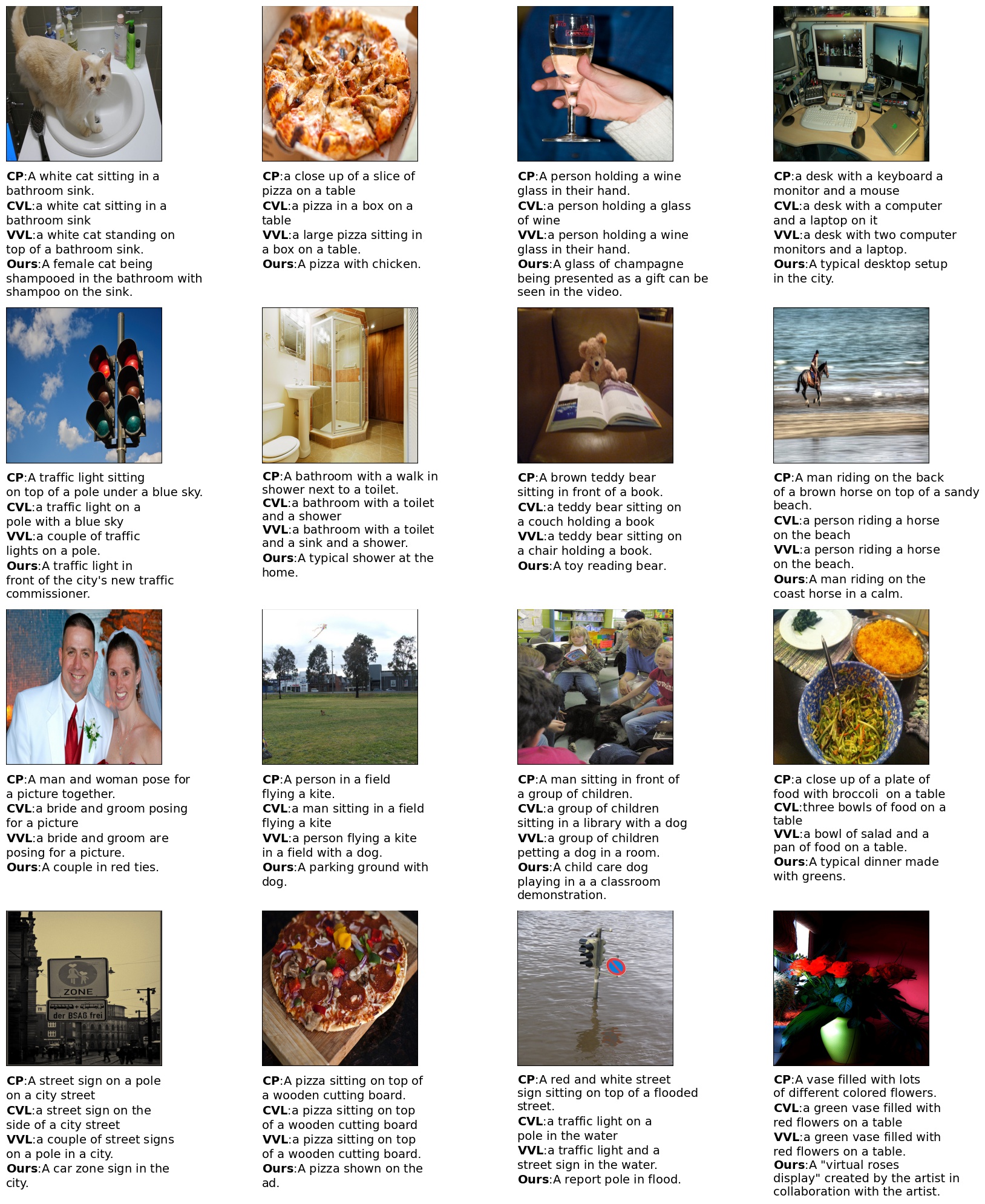}
\end{figure*}
\begin{figure*}[h!]
	\centering
    \ContinuedFloat
    \includegraphics[width=0.95\linewidth]{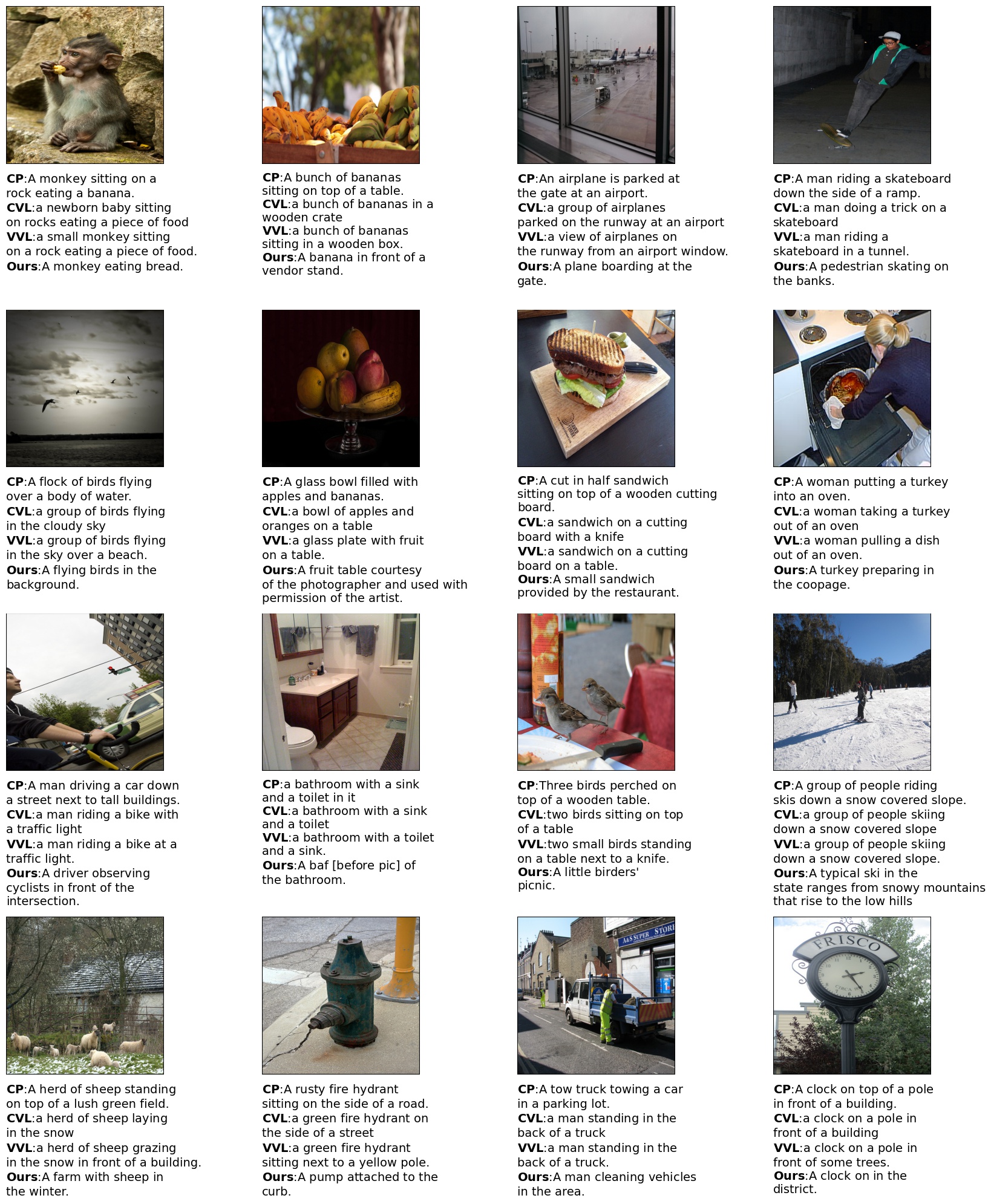}
\end{figure*}
\begin{figure*}[h!]
	\centering
    \ContinuedFloat
    \includegraphics[width=0.95\linewidth]{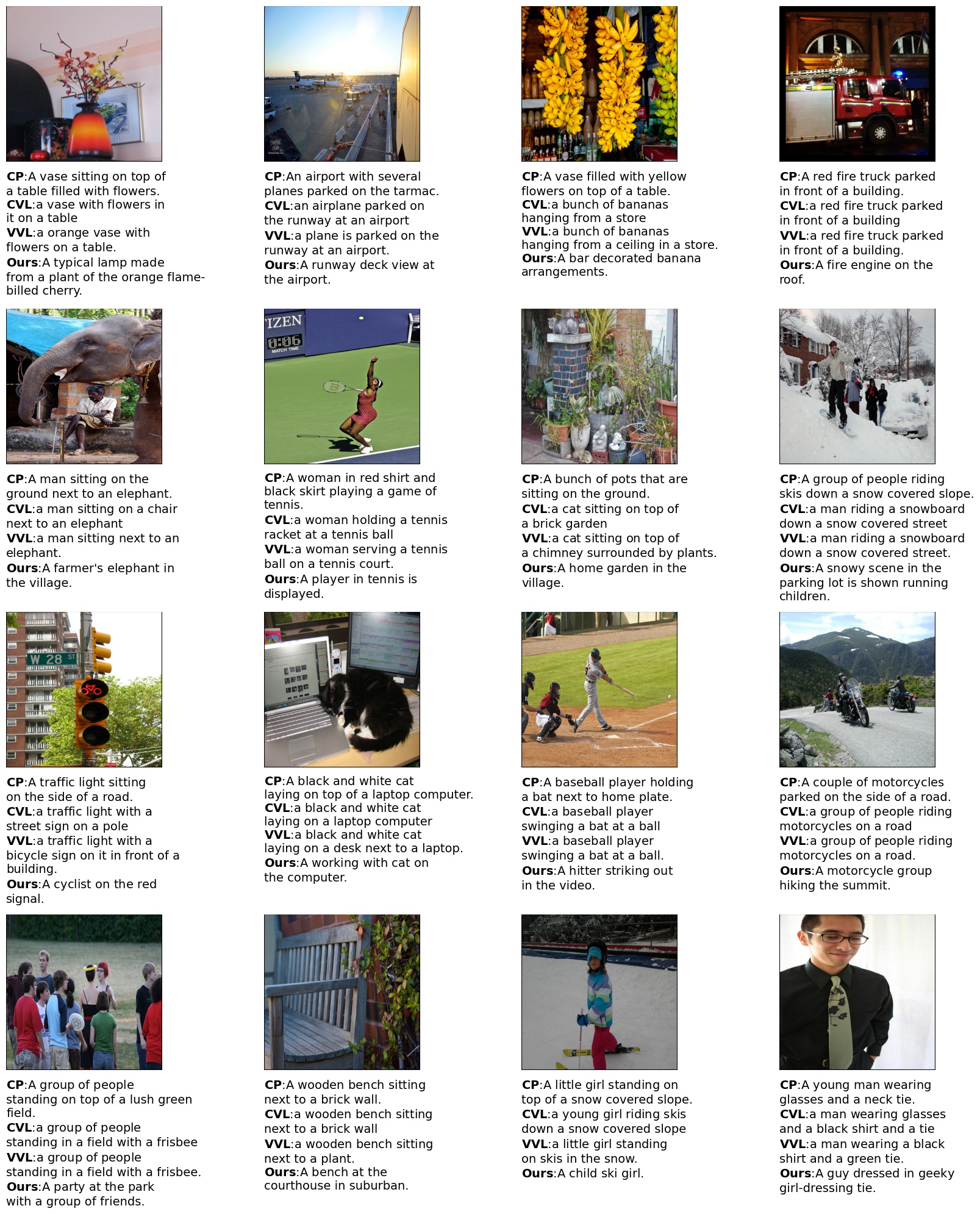}
\end{figure*}
\begin{figure*}[h!]
	\centering
    \ContinuedFloat
    \includegraphics[width=0.95\linewidth]{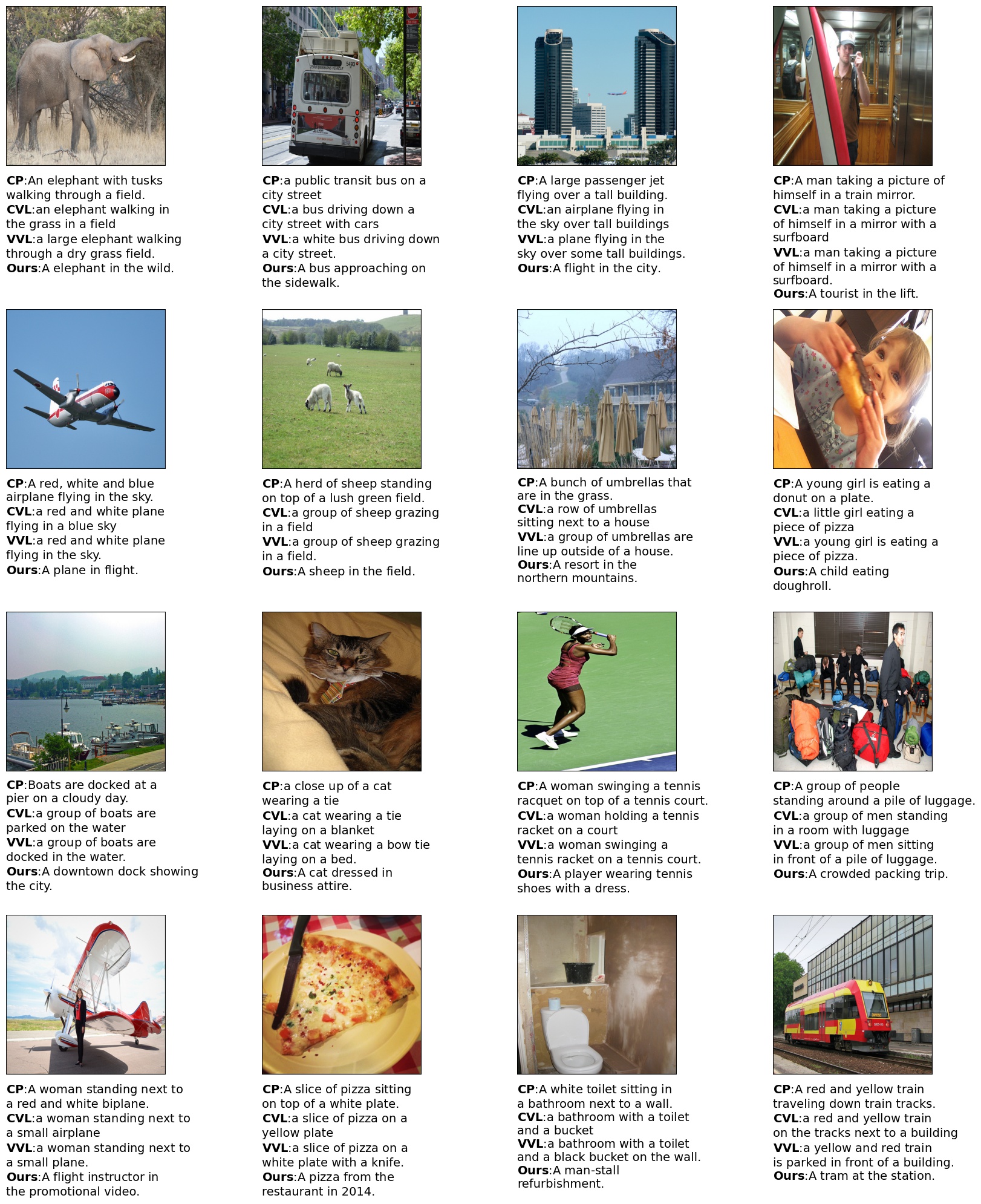}
\end{figure*}
\begin{figure*}[h!]
	\centering
    \ContinuedFloat
    \includegraphics[width=0.95\linewidth]{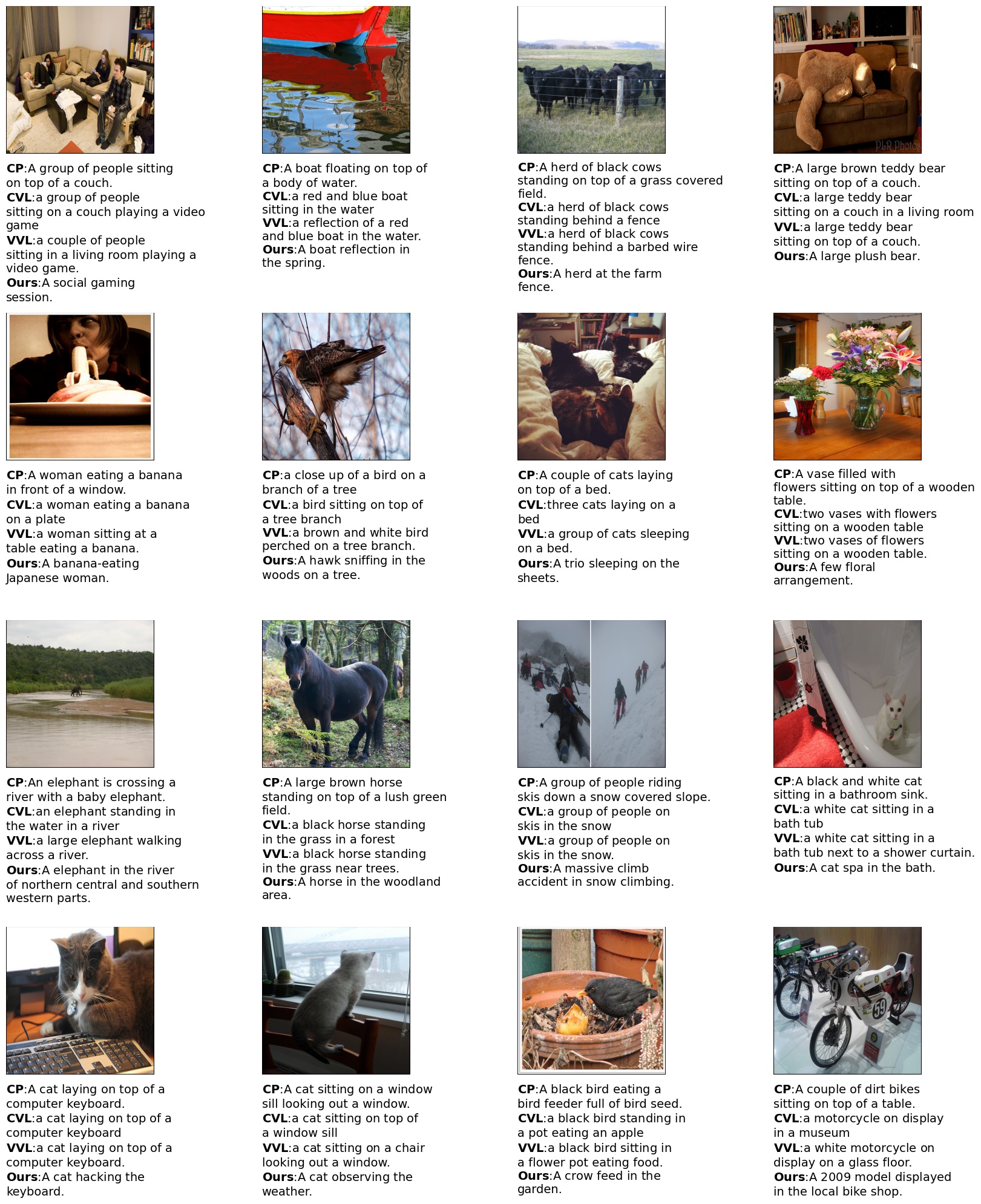}
\end{figure*}

\begin{figure*}[h!]
	\centering
    \ContinuedFloat
    \includegraphics[width=0.95\linewidth]{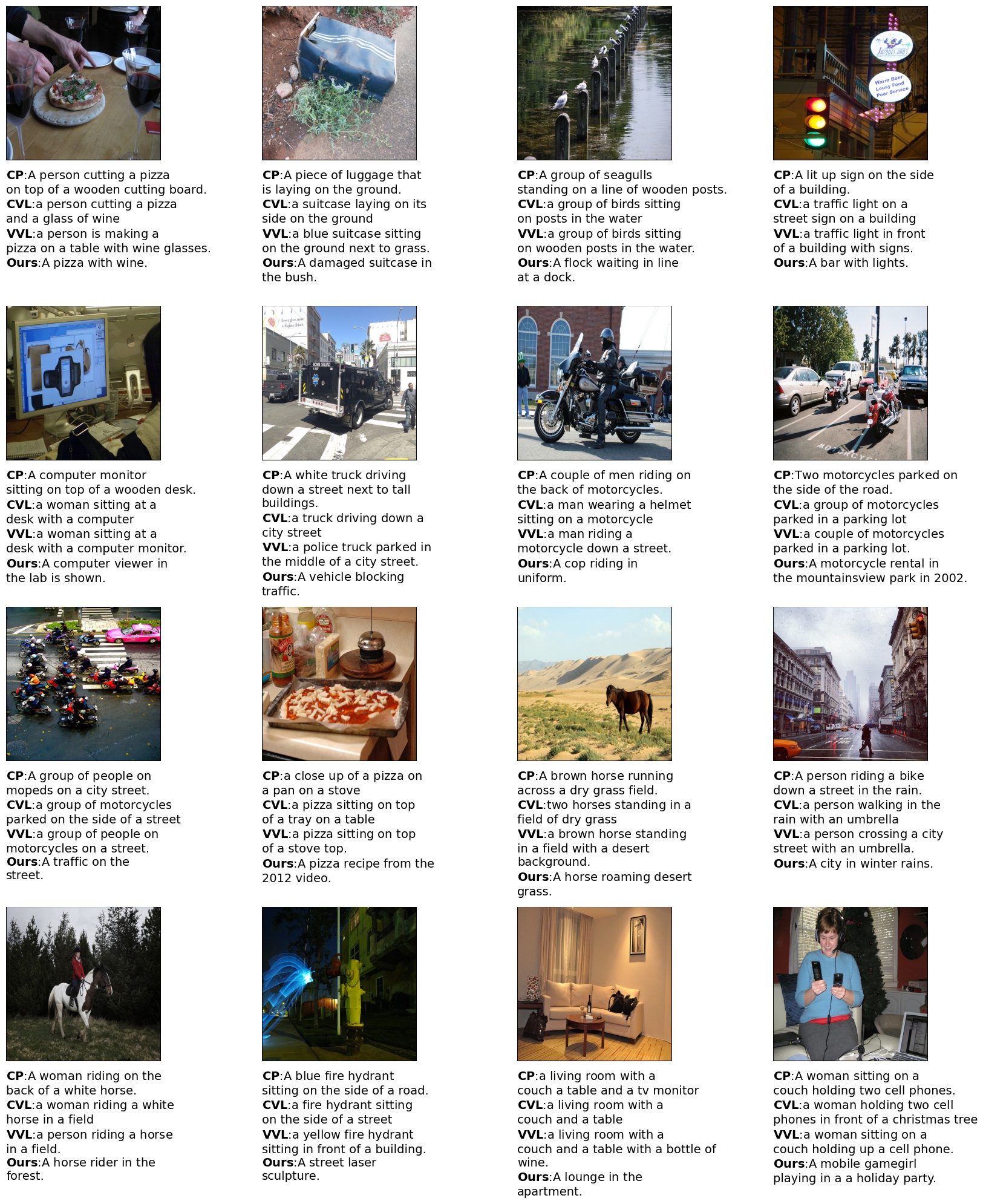}
\end{figure*}


\begin{figure*}[h!]
	\centering
    \ContinuedFloat
    \includegraphics[width=0.95\linewidth]{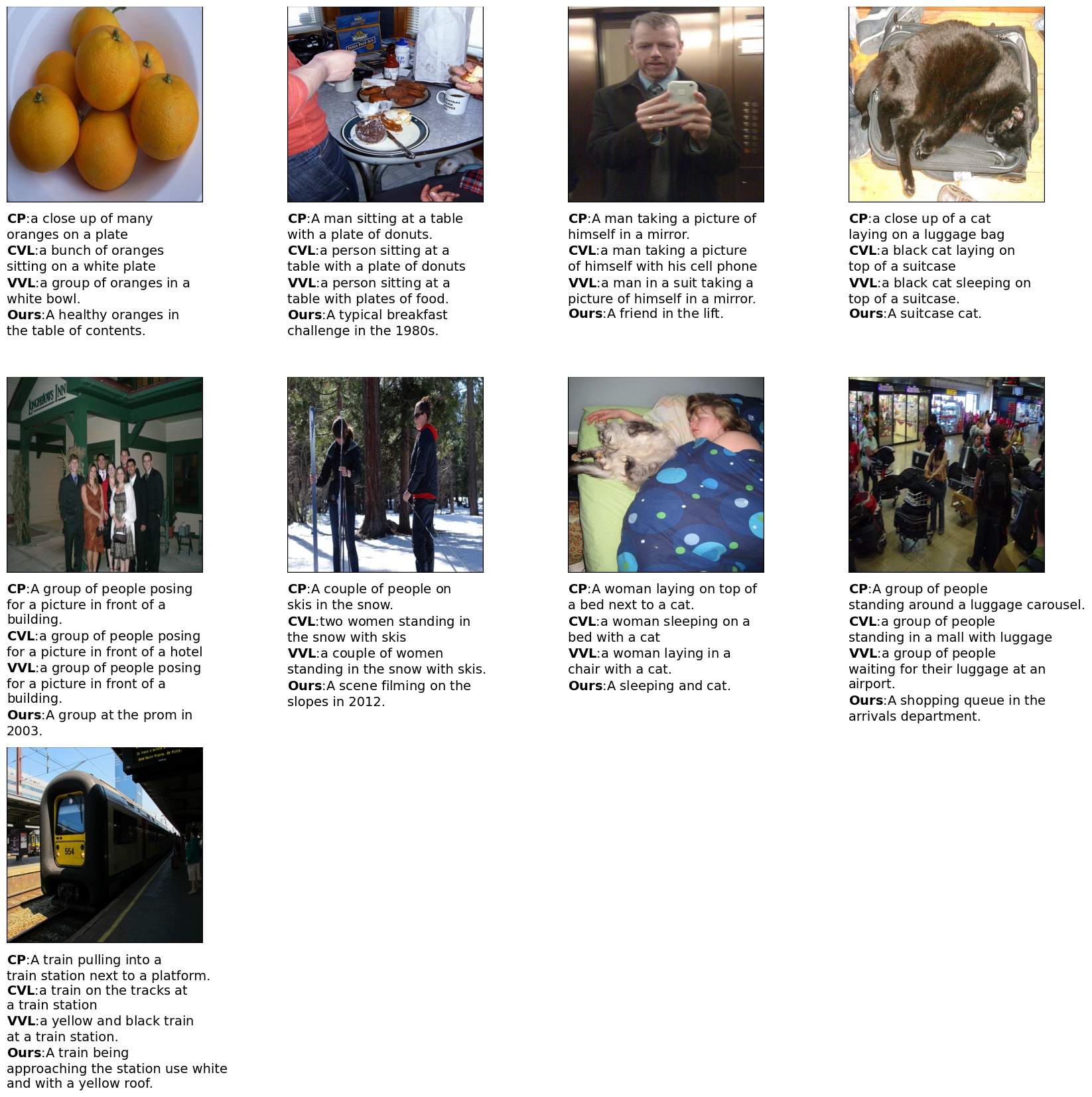}
\end{figure*}

\end{document}